\documentclass{article}

\usepackage{microtype}
\usepackage{graphicx}
\usepackage{subcaption}
\usepackage{booktabs} % for professional tables
\usepackage{wrapfig}
\usepackage{hyperref}
\usepackage[most]{tcolorbox}
\usepackage{multirow} 
\usepackage[preprint]{icml2026}

\usepackage{enumitem}

\usepackage{amsmath}
\usepackage{amssymb}
\usepackage{mathtools}
\usepackage{amsthm}
\usepackage[subrefformat=parens]{subcaption}
\usepackage[capitalize,noabbrev]{cleveref}

\theoremstyle{plain}
\newtheorem{theorem}{Theorem}

\theoremstyle{definition}
\newtheorem{definition}[theorem]{Definition}

\theoremstyle{remark}

\usepackage[textsize=tiny]{todonotes}

\icmltitlerunning{Preprint under review}

\begin{document}

\twocolumn[
  % \icmltitle{Flexible Feature Distillation for Large Language Models}
    % \icmltitle{Selective Subspace Alignment for Efficient Feature Distillation in Large Language Models}
    \icmltitle{What Should Feature Distillation Transfer in LLMs? \\ A Task-Tangent Geometry View}



  \icmlsetsymbol{equal}{*}

\author{%
  Khouloud Saadi\thanks{Corresponding author.} \\
  Department of Computer Science\\
  KAUST, Thuwal, Saudi Arabia \\
  \texttt{khouloud.saadi@kaust.edu.sa} \\
  % examples of more authors
  \And
  Di Wang \\
  Department of Computer Science \\
  KAUST, Thuwal, Saudi Arabia \\
  \texttt{di.wang@kaust.edu.sa} \\
  }
    \begin{icmlauthorlist}
    \icmlauthor{Khouloud Saadi}{yyy} %{equal,yyy}
    \icmlauthor{Di Wang}{yyy} %{equal,yyy,comp}
    % \icmlauthor{Firstname3 Lastname3}{comp}
  \end{icmlauthorlist}

  \icmlaffiliation{yyy}{Department of Computer Science, King Abdullah University of Science and Technology, Thuwal, Saudi Arabia}

  \icmlcorrespondingauthor{Khouloud Saadi}{Khouloud.saadi@kaust.edu.sa}
  
  % \begin{icmlauthorlist}
  %   \icmlauthor{Firstname1 Lastname1}{equal,yyy}
  %   \icmlauthor{Firstname2 Lastname2}{equal,yyy,comp}
  %   \icmlauthor{Firstname3 Lastname3}{comp}
  %   \icmlauthor{Firstname4 Lastname4}{sch}
  %   \icmlauthor{Firstname5 Lastname5}{yyy}
  %   \icmlauthor{Firstname6 Lastname6}{sch,yyy,comp}
  %   \icmlauthor{Firstname7 Lastname7}{comp}
  %   %\icmlauthor{}{sch}
  %   \icmlauthor{Firstname8 Lastname8}{sch}
  %   \icmlauthor{Firstname8 Lastname8}{yyy,comp}
  %   %\icmlauthor{}{sch}
  %   %\icmlauthor{}{sch}
  % \end{icmlauthorlist}



  \icmlkeywords{Machine Learning, ICML}

  \vskip 0.3in
]



\printAffiliationsAndNotice{}  % no special notice (required even if empty)

\begin{abstract}
Feature-based knowledge distillation aims to transfer intermediate representations from a teacher LLM model to a student. Existing approaches typically rely on direct feature matching or learned projections, implicitly treating representations as objects with intrinsic meaning. However, the relevance of a representation dimension is determined solely by how it affects the model’s output. In this work, we propose a functional perspective on feature-based distillation. We characterize knowledge transfer in terms of the teacher’s functional geometry, i.e., how its output depends on internal representations, rather than direct representation alignment. This viewpoint reveals that effective distillation need not preserve full high-dimensional features, but instead should retain dominant directions of functional contribution, naturally inducing an effective functional dimension for each task. Building on this framework, we introduce \emph{Flex-KD}, an architecture-agnostic and parameter-free distillation method that transfers the teacher’s functional geometry while matching the student’s representational capacity. Extensive experiments across language understanding and generation benchmarks demonstrate that Flex-KD consistently outperforms existing distillation approaches, particularly under severe teacher–student dimension mismatch.
\end{abstract}

\section{Introduction}

Large language models (LLMs) have achieved remarkable performance across a wide range of classification and generation tasks \citep{liangmixkd,jiao-etal-2020-tinybert,liu2024deepseek,openai2023gpt,team2023gemini}. However, their substantial computational and memory costs often limit practical deployment. Knowledge distillation (KD) has therefore become a central approach for transferring the capabilities of large models into smaller, task-specific students \citep{schmidhuber1992learning,hinton2015distilling,zhu2024survey,xu2022survey}, particularly during fine-tuning on downstream tasks \citep{zhou2021bert,liang2020mixkd,Sun2019PatientKD,gu2024minillm,kodistillm}. Most existing LLM distillation methods focus on logit-based objectives, which transfer softened output distributions from teacher to student \citep{gu2024minillm,taori2023stanford,kim-rush-2016-sequence}. Feature-based distillation, which leverages intermediate representations, offers a complementary signal and has shown strong empirical results \citep{sanh2019distilbert,sun-etal-2019-patient,dasguptaimproving,saadi2023learn2}, yet remains comparatively underexplored in modern LLM settings.

A fundamental limitation of existing feature KD approaches lies in how representations are treated. Most methods aim to directly align teacher and student hidden states, either through matching losses or learned projection layers \citep{jiao-etal-2020-tinybert,chen2022improved}. This formulation implicitly assumes that representations possess intrinsic meaning. In practice, however, internal representations are intermediate coordinates whose relevance is entirely determined by how they affect the model’s output. As a result, matching representations without accounting for their functional role can be inefficient, particularly when the teacher and student differ substantially in representational capacity.

This issue is exacerbated in LLM distillation, where students are often orders of magnitude smaller than their teachers. Projector-based alignment introduces additional parameters that must be learned from limited downstream data and may distort task-relevant structure~\citep{dasguptaimproving,miles2024understanding}. More fundamentally, transferring full high-dimensional representations ignores the fact that only a subset of representation components contributes meaningfully to a given task. Recent studies have shown that many LLM units play a limited role in downstream performance, while functional contributions are concentrated in a small subset of components~\citep{hase2024does,gromov2024unreasonable,luo2024sparsing,dai2021knowledge}.

Despite their empirical success, existing LLMs feature-based KD~\citep{dasguptaimproving,miles2024understanding} methods implicitly rely on a strong but rarely stated assumption: that intermediate representations possess intrinsic semantic meaning, and that aligning them across teacher and student models is therefore desirable. However, intermediate representations are not objectives in themselves; their only relevance lies in how they influence the teacher’s input–output function. From this perspective, directly matching representations, whether through projection, regression, or similarity-based objectives, addresses a proxy problem that may not reflect what truly needs to be transferred.

In this work, we argue that feature distillation should instead be formulated in functional terms. Rather than asking how to align representations, we ask which directions in the teacher’s representation space are functionally relevant, in the sense that perturbations along those directions induce meaningful changes in the teacher’s output. This shift reframes feature distillation as the problem of preserving the teacher’s \emph{function-induced geometry} on its hidden representations, providing a principled criterion for determining what information should be transferred.

Building on this insights, we propose \textbf{Flex-KD}, a parameter-free method that instantiates functional geometry transfer in practice. Flex-KD identifies task-relevant representation directions via output contributions, constructs a functionally meaningful subspace matched to the student’s capacity, and performs geometry-preserving distillation through correlation-based alignment. This design yields a distillation objective that remains well-defined under severe teacher--student dimension mismatch, without introducing additional projection parameters.A detailed \textbf{related work} section is in Appendix~\ref{sec:RL}. 

To sum up, our contributions are as follow: 
\begin{itemize}[leftmargin=*, nosep]
    \item We revisit feature KD in LLMs from a functional perspective, arguing that intermediate representations are only meaningful insofar as they influence model outputs, and that effective transfer requires preserving task-dependent functional geometry rather than raw feature similarity.
    \item We formalize this view by introducing the notion of an effective functional dimension, showing that a model’s output variation is often concentrated in a small set of representation directions and motivating a principle of functional geometry preservation.
    \item We propose \textbf{Flex-KD}, a parameter-free and dimension-aware LLMs distillation method that aligns teacher and student models within a task-relevant functional subspace, naturally supporting flexible teacher–student dimension mismatch without learned projectors.
    \item Extensive experiments across classification and generation tasks demonstrate consistent improvements over feature KD baselines, particularly under strong compression, representation mismatch, and limited-data regimes.
\end{itemize}
\vspace{-1.8 em}
\section{A Functional View of Feature-Based Distillation}
\label{sec:theory}

In this section, we formalize the functional perspective introduced above by characterizing how the teacher’s LLM input–output function induces a geometry over its hidden representations. Our analysis departs from conventional representation-centric views, which study hidden states in LLMs as geometric objects in their own right. Instead, we consider representations only through the lens of the function they support. Specifically, we study the local sensitivity of the teacher’s output with respect to perturbations in its hidden representation. This induces a function-tangent geometry, in which directions are not defined by representation similarity, but by their influence on the teacher’s behavior. Under this view, two representation directions are equivalent if they induce comparable functional effects, regardless of their coordinate values or semantic interpretability. This functional geometry provides the foundation for determining which components of a teacher-LLM representation are essential to preserve during distillation.

\vspace{-1em}
\subsection{Local Functional Dependence}

To make this dependence explicit, consider the local behavior of the teacher model around the representation induced by an input $x$. Let $f_T(x)$ denote the teacher output and let $h^T(x) \in \mathbb{R}^{d_T}$ be the hidden representation at a chosen layer. For small perturbations $\delta$ applied to this representation, the resulting change in the output can be approximated as
\begin{equation}
f_T(h^T(x) + \delta)
\;\approx\;
f_T(h^T(x)) + J_T(x)\,\delta,
\label{eq:linearization}
\end{equation}
where
$
J_T(x) = \frac{\partial f_T(x)}{\partial h^T}
$
denotes the Jacobian of the teacher output with respect to its hidden representation.

This local linearization is not an assumption about the global structure of the network, but rather an interpretive lens. It highlights which representation directions locally contribute to the teacher’s output. Perturbations along directions associated with large Jacobian magnitude induce significant output variation, while perturbations along directions with small magnitude have negligible functional effect. Consequently, the functional role of a representation dimension can be characterized by the contribution of the output with respect to that dimension.
\vspace{-1em}

\subsection{Function-Tangent Geometry} \label{FTG}

Motivated by the local dependence in Eq.~\eqref{eq:linearization}, we characterize functional relevance through output contribution.

\begin{definition}[Function-tangent vector]
For a given input $x$, we define the function-tangent vector as:
\begin{equation}
g(x) = \nabla_{h^T} f_T(x),\footnote{
For models with vector-valued outputs, such as classification logits or next-token distributions in LLMs, $f_T(x)$ denotes a scalar functional of the output. In practice, we aggregate gradients across output coordinates by computing the gradient of the summed log-probabilities of the teacher output distribution with respect to the hidden representation, yielding a single contribution score per representation dimension.
}
\end{equation}
which measures how the teacher’s output varies with respect to perturbations of each representation coordinate.
\end{definition}

\vspace{-1.8 em}
Each coordinate of $g(x)$ quantifies the local functional contributions of the corresponding representation direction. Directions with large magnitude strongly affect the teacher’s output, whereas directions with small magnitude have limited functional impact. Function-tangent vectors therefore describe the \emph{local functional geometry} of the teacher model in representation space.

\textbf{Aggregated Functional Contribution.} While $g(x)$ captures functional contribution for a single input, distillation requires identifying representation directions that are consistently relevant across the dataset. To this end, we aggregate function-tangent magnitudes over the data distribution: $
G = \mathbb{E}_{x}\left[\,\left| \nabla_{h^T} f_T(x) \right|\,\right] \in \mathbb{R}^{d_T}$, is the overall functional contribution of each representation dimension, where the absolute value is applied element-wise. 

\vspace{-1em}
\subsection{Effective Functional Dimension}

Large language models are heavily overparameterized \citep{liu2024deepseek,openai2023gpt,team2023gemini}, with representation dimensionalities that often far exceed what is required for a specific downstream task. Recent empirical analyses of LLM internals
\citep{hase2024does,gromov2024unreasonable,luo2024sparsing,dai2021knowledge}
have consistently shown that task performance is driven by a limited subset of components, while many representation dimensions exhibit negligible functional contribution. This behavior is especially pronounced in final-layer representations, where task predictions are formed~\citep{gromov2024unreasonable, men2024shortgpt}. Motivated by these observations, we formalize the notion of an \emph{effective functional dimension}, capturing how many representation directions are meaningfully involved in determining the teacher’s output behavior.
Let $G_{(1)} \ge \cdots \ge G_{(d_T)}$ denote the sorted entries of $G$. For a target dimension $k$, we define the functional tail mass:
\setlength{\abovedisplayskip}{4pt}
\setlength{\belowdisplayskip}{4pt}
\begin{equation}
\mathrm{Tail}(k)=\sum_{i>k} G_{(i)},
\end{equation}
\setlength{\abovedisplayskip}{6pt} 
\setlength{\belowdisplayskip}{6pt}
A small tail mass indicates that the teacher’s output behavior is dominated by a small number of representation coordinates. Accordingly, we interpret the \emph{effective functional dimension} as the smallest $k$ such that $\mathrm{Tail}(k)$ becomes negligible, yielding a task-dependent measure of how many representation directions are needed to capture the teacher’s functional behavior.
\vspace{-1.0em}
\subsection{Principle of Functional Geometry Preservation}
The above formulation leads to a fundamental principle for feature-based distillation:
\vspace{-0.6em}
\begin{tcolorbox}[
  colback=blue!2,
  colframe=blue!40!black,
  title=\textbf{Principle: Functional Geometry Preservation},
  center title,
  boxsep=2pt,
  top=2pt,
  bottom=2pt
]
\emph{
The student’s ability to approximate the teacher’s local output behavior
depends on preserving the teacher’s dominant function-tangent directions.
If distillation is restricted to a $k$-dimensional subspace that captures most
of the aggregated functional contribution, the essential functional geometry is
retained; directions with small tail mass $\mathrm{Tail}(k)$ have limited
influence on output behavior.
}
\end{tcolorbox}
This principle suggests that effective feature-based KD need not transfer the full teacher representation: it is sufficient to preserve the dominant function-tangent directions while deemphasizing coordinates with negligible contribution.

\begin{figure*}[t]
    \centering
    \includegraphics[width=0.99\linewidth]{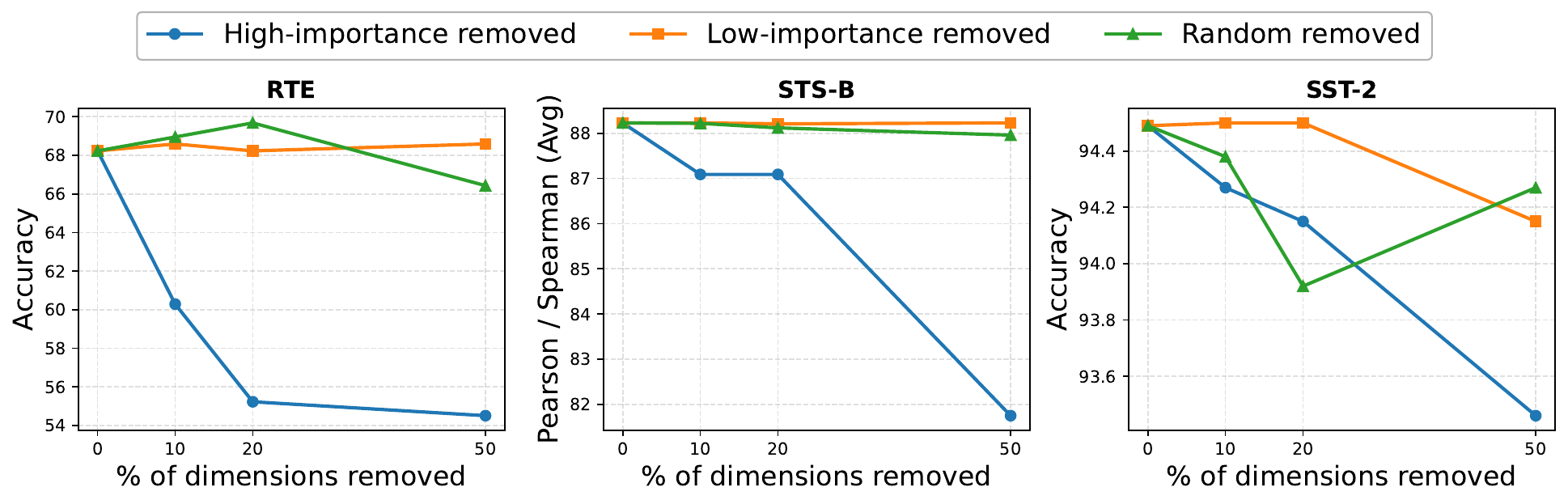}
    \caption{
    \textbf{Functional importance of representation dimensions.}
    Performance after selectively removing teacher representation coordinates ranked by estimated functional contribution.
    Removing highly important functional directions leads to rapid degradation across tasks, whereas removing low-importance directions has negligible effect even when up to 50\% of dimensions are removed.
    Random removal exhibits intermediate behavior.
    These results provide direct empirical evidence that learned representations possess a highly uneven functional structure, supporting the notion of a low effective functional dimension.
    }
    \label{fig:functional_importance}
    \vspace{-1.5em}
\end{figure*}

To assess whether the proposed functional importance measure captures meaningful structure in learned representations, we conduct a controlled ablation study in which teacher representation coordinates are selectively removed according to their estimated contribution. Specifically, we compare removing the most important coordinates, the least important coordinates, and randomly selected coordinates, and evaluate the resulting downstream task performance. As shown in Figure~\ref{fig:functional_importance}, a consistent and clear pattern emerges across RTE, STS-B, and SST-2. Removing highly important directions leads to rapid and substantial performance degradation, even at modest removal rates. In contrast, removing low-importance directions has little to no effect, even when up to $50\%$ of representation dimensions are discarded, while random removal exhibits intermediate behavior. These results provide direct empirical evidence that teacher representations possess a highly uneven functional structure, and that the proposed importance measure reliably identifies the directions that govern task behavior.

\vspace{-1.0em}

\subsection{Consequences for Feature KD Objectives}

The functional geometry framework reveals that representation dimensions differ substantially in their influence on the teacher’s output behavior. This heterogeneity is captured by the aggregated functional contribution vector $G$, whose entries quantify the expected sensitivity of the teacher output to local perturbations of individual representation coordinates.
To make this notion precise, consider a small perturbation $\delta \in \mathbb{R}^{d_T}$ applied to the teacher representation.
A first-order expansion around $h^T(x)$ yields the following functional decomposition:
\setlength{\abovedisplayskip}{6pt}  % restore defaults
\setlength{\belowdisplayskip}{6pt}
\begin{equation}
\mathbb{E}_x
\left[
\left| f_T(h^T(x)+\delta) - f_T(h^T(x)) \right|
\right]
\le
\sum_{i=1}^{d_T} G_i\,|\delta_i|
+
\mathcal{O}(\|\delta\|_2^2),
\label{eq:coord_bound_1}
\end{equation}
\setlength{\abovedisplayskip}{6pt}  % restore defaults
\setlength{\belowdisplayskip}{6pt}
which characterizes how perturbations along different representation directions $\delta_i$ contribute unevenly (with a corresponding scale $G_i$) to changes in the teacher’s output.
Intuitively, if a small perturbation along a given direction induces a large change in the output, then that direction plays a central role in the model’s computation. Hence, directions along which perturbations have little effect contribute minimally to task behavior. The resulting decomposition therefore provides a principled way to distinguish dominant and negligible directions based on their functional influence. Next, let $E_k$ denote the index set of the top-$k$ representation coordinates ranked by functional contribution. Restricting attention to this dominant subset yields:
\begin{align}
\mathbb{E}_x
\left[
\left| f_T(h^T(x)+\delta) - f_T(h^T(x)) \right|
\right]
& \le
 \sum_{i \in E_k} G_i\,|\delta_i| \nonumber \\
 + & \mathrm{Tail}(k,\delta)
+ \mathcal{O}(\|\delta\|_2^2),
\label{eq:tail_bound}
\end{align}
where
$
\mathrm{Tail}(k,\delta) = \sum_{i \notin E_k} G_i\,|\delta_i|
$
quantifies the cumulative functional contribution of discarded directions. The full proof is available in Appendix~\ref{app:coord_proof}.

Equations~\eqref{eq:coord_bound_1}–\eqref{eq:tail_bound} provide a local characterization of effective functional dimension.
When the functional tail mass is small, variations outside the dominant subspace induce only limited changes in the teacher’s output behavior, indicating that task-relevant computation is concentrated in a low-dimensional functional subspace.
Crucially, this quantity reflects a structural property of the learned representation, rather than a guarantee enforced by training.

Importantly, while Equation~\eqref{eq:tail_bound} does not assume the functional tail to be small \emph{a priori}, empirical evidence suggests that functional contributions are often highly concentrated in practice.
As shown in Figure~\ref{fig:functional_importance}, removing a substantial fraction of the least important directions, ranked according to the proposed functional importance measure, typically results in little to no degradation in downstream performance, whereas removing high-importance directions leads to rapid failure. This observation supports the relevance of selective subspace transfer for real-world models and tasks.

This perspective clarifies a limitation of feature KD objectives defined over the full representation space.
Projection-based methods~\citep{jiao-etal-2020-tinybert,chen2022improved} allocate modeling capacity across all representation coordinates, including those with negligible functional contribution.
Similarly, covariance-based formulations such as CKA~\citep{dasguptaimproving} are agnostic to functional contribution as captured by $G$ treat all coordinates symmetrically, regardless of their influence on the teacher’s output behavior.
When functional contributions are highly uneven, full-space alignment may therefore expend optimization effort on directions that are largely irrelevant for task behavior.

Finally, the functional geometry view suggests why selective transfer can be particularly beneficial in low-data regimes. When supervision is limited, learning to suppress functionally irrelevant directions through optimization alone can be statistically inefficient.
Restricting distillation to directions with high functional contribution introduces a task-informed inductive bias that focuses learning on the subspace most directly governing output behavior.

\vspace{-1em}
\section{Flex-KD: Functional Geometry Transfer}
\label{sec:method}
Flex-KD is a direct instantiation of Section~\ref{sec:theory}’s functional-geometry principle. We first estimate per-coordinate output sensitivity and aggregate it across data to approximate $G$. Then, select the top-$d_s$ coordinates to form a student-sized functional subspace (corresponding to choosing $k=d_s$ in the tail-mass view). Distillation is then performed by aligning teacher and student representations only within this subspace, using a correlation objective that preserves directional structure while remaining invariant to scale. Because the objective is defined on a selected subspace rather than the full teacher width, it remains well-posed under severe teacher–student dimension mismatch without introducing learned projectors.

\vspace{-1.3em}
\subsection{Estimating Function-Tangent Directions}

Let $h^T(x) \in \mathbb{R}^{d_T}$ denote the teacher’s hidden representation at the selected layer. To characterize how the teacher’s output depends on this representation, we estimate functional contribution by computing gradients of the output with respect to $h^T(x)$. For each input sample $x_j$:
\begin{equation}
g(x_j) = \left| \frac{\partial f_T(x_j)}{\partial h^T} \right| \in \mathbb{R}^{d_T},
\end{equation}
where the absolute value is applied element-wise. Each coordinate of $g(x_j)$ quantifies how strongly the corresponding representation direction contributes to variations in the teacher’s output. This provides a first-order characterization of the teacher’s local functional geometry in representation space, consistent with the framework developed in Section~\ref{sec:theory}.
\vspace{-2em}
\subsection{Constructing a Functionally Relevant Subspace}
While functional contribution is input-dependent, distillation requires identifying representation directions that are consistently relevant across the data distribution. As introduced in Section~\ref{sec:theory}, this corresponds to the population-level functional contribution vector
$
G = \mathbb{E}_x \left[\,|\nabla_{h^T} f_T(x)|\,\right].
$
In practice, this quantity is unknown and must be estimated from data. We therefore construct an empirical estimator by aggregating gradient magnitudes over the dataset:
\setlength{\abovedisplayskip}{6pt}  % restore defaults
\setlength{\belowdisplayskip}{6pt}
\begin{equation}
\hat{G} = \frac{1}{N} \sum_{j=1}^{N} g(x_j) \in \mathbb{R}^{d_T},
\end{equation}
\setlength{\abovedisplayskip}{6pt}  % restore defaults
\setlength{\belowdisplayskip}{6pt}
% \vspace{-0.5em}
The estimator $\hat{G}$ provides a practical approximation of the functional geometry introduced in Section~\ref{sec:theory}. We rank representation dimensions according to $\hat{G}$ and select the top-$d_S$ indices
$
E = \{i_1, i_2, \ldots, i_{d_S}\},
$
thereby constructing a $d_S$-dimensional subspace that approximates the dominant components of the teacher’s functional geometry.
\vspace{-1.0em}
\subsection{Geometry-Preserving Distillation}

Given the selected index set $E$, we extract the corresponding teacher representation
$h^{T_{d_S}} \in \mathbb{R}^{d_S}$.
Functional importance identifies the representation directions that govern local input–output behavior; however, an additional question concerns how these directions should be transferred across heterogeneous teacher–student architectures.
Representation coordinates do not possess intrinsic meaning: their scale and parameterization are model-dependent and therefore arbitrary.
As a result, pointwise feature matching objectives such as MSE or cosine distance may impose constraints that are misaligned with functional geometry.
To preserve task-relevant local behavior while remaining invariant to per-dimension scaling, we adopt a correlation-based alignment loss. An ablation (Table~6 in Appendix~\ref{loss}) confirms that correlation-based alignment consistently outperforms MSE and cosine alternatives, motivating the formulation in Eq.~\ref{corr_loss}:
\vspace{-0.1em}
\begin{equation} \label{corr_loss}
L_{\text{Flex-KD}} = \sum_{m=1}^{d_S} (1 - C_{mm})^2 ,
\end{equation}
where $C_{mm}$ denotes the correlation between the $m$-th student feature and its corresponding selected teacher directions. To simplify notations, $h^{T_{d_{S}}}$ and $h^{S}$ are assumed to be mean-centered along the batch dimension, such that each unit has mean output $0$ over the batch $n$.
% For a batch of $n$ samples, the cross-correlation is defined as
\begin{equation}
C_{mm} =
\frac{\sum_{j=1}^{n} h^{T_{d_S}}_{j,i_m} \, h^S_{j,m}}
{\sqrt{\sum_{j=1}^{n} (h^{T_{d_S}}_{j,i_m})^2}
 \sqrt{\sum_{j=1}^{n} (h^S_{j,m})^2}} ,
\end{equation}

This objective preserves the orientation of teacher and student representations along functionally dominant directions while remaining invariant to feature rescaling, making it well aligned with functional geometry transfer.
The final training objective is given by:
\begin{equation}
L_{\text{final}} =
\alpha L_{\text{Flex-KD}} +
\beta L_{\text{logit}} +
\lambda L_{\text{sup}},
\end{equation}
where $L_{\text{logit}}$ denotes standard logit-based distillation and $L_{\text{sup}}$ is the supervised task loss.
Flex-KD can be applied either independently or in combination with logit distillation.
% \vspace{-0.3em}
% \subsection{Discussion}
Flex-KD provides a concrete instantiation of functional geometry transfer for feature-based knowledge distillation. By explicitly controlling \emph{what} information is transferred, \emph{how much} information is transferred, and \emph{which geometric structure} is preserved, the method departs from conventional projector-based formulations that attempt to align full representations. This perspective yields a distillation objective that remains well-defined under severe teacher--student dimension mismatch, while remaining simple, parameter-free, and stable in practice.

\vspace{-1em}

\section{Experimental Results }
We evaluate our approach across three core tasks: text classification, instruction-following, and summarization. Following \citet{dasguptaimproving, jiao-etal-2020-tinybert, sanh2019distilbert}, all teacher models are static during distillation. Details on the aggregation procedure used to estimate the most important directions in the teacher space are provided in Appendix~\ref{aggregation}.

\begin{table}[!t]
\centering

\caption{
Test accuracy (\%) on the IMDB dataset, averaged over three random seeds. Values in \textcolor{green!60!black}{green} denote gains over the KD baseline, while values in \textcolor{red}{red} indicate drops. For GPT2, distillation is from $h^{T}=1024$ to $h^{S}=768$; for BERT, from $h^{T}=768$ to $h^{S}=312$.} 
\resizebox{0.95\linewidth}{!}{%
\begin{tabular}{@{}lcc@{}}
\toprule
\multirow{2}{*}{Method} & \multicolumn{1}{c}{345M $\rightarrow$ 124M} & \multicolumn{1}{c}{110M $\rightarrow$ 14M} \\
                        & GPT2 & BERT \\
\midrule
Teacher         & 95.47         & 94.06         \\
\cmidrule(r){1-3}
FT \citeyearpar{devlin2019bert}                 & $94.20\pm0.30 
 $        & $89.24\pm0.08$ \\
KD \citeyearpar{hinton2015distilling}                  & $94.21\pm0.42$         & $89.58\pm0.10$ \\
\textit{Projector} \citeyearpar{jiao-etal-2020-tinybert}                  & $94.01\pm0.12$ {\color{red} \small (-0.20)} & $89.39\pm0.05$ {\color{red} \small (-0.19)} \\
\textit{CKA} \citeyearpar{dasguptaimproving}                  & $\underline{94.65\pm0.10}$ {\color{green!60!black} \small (+0.44)}& $\underline{90.13\pm0.06}$ {\color{green!60!black} \small (+0.55)}\\
\textbf{\textit{Flex-KD}}         & \textbf{\boldmath$95.09\pm0.04$} {\color{green!60!black} \small (+0.88)} & \textbf{\boldmath$90.60\pm0.04$} {\color{green!60!black} \small (+1.02)}\\
\bottomrule
\end{tabular}}
\label{table:one}
\vspace{-1.9em}
\end{table}

\vspace{-1 em}
\subsection{Classification} \label{classifi}

We first evaluate Flex-KD on sentence-level classification tasks, using IMDB \citep{maas2011learning} and five GLUE tasks \citep{wang-etal-2018-glue}. This setting directly tests the central claim of Section~\ref{sec:theory}: when teacher and student have mismatched hidden dimensions, distillation should prioritize the teacher directions that most affect the output, rather than matching representations indiscriminately. The full training details are provided in Subsection~\ref{expdetails} of Appendix~\ref{exp}.

\begin{table*}[t]
    \centering
    \caption{Results (in \%) averaged over three random seeds. The teacher model is GPT2-medium (345M parameters) and the student model is GPT2-small (124M parameters). “AVG” denotes the average performance across all tasks. Values in \textcolor{green!60!black}{green} indicate performance gains over the KD baseline, while those in \textcolor{red}{red} indicate performance drops. Feature distillation is performed from $h^{T}=1024$ to $h^{S}=768$. For the full table with standard deviations, see Table~\ref{table:six} in Appendix~\ref{perglue}.}
  % \resizebox{0.95\linewidth}{!}{%
    \begin{tabular}{@{}lcccccc@{}}
        \toprule
 Method &  SST-2 & STS-B & MRPC & RTE&MNLI&AVG\\ \midrule   
    Teacher(12 x 1024)&94.49 &	88.23&  84.09& 68.23&85.10   & 84.02\\
        \cmidrule(r){1-7}

FT \citep{devlin2019bert}   & $91.32$ &  $86.58$& $81.68$&\boldmath$65.95$ & $81.78$&81.46\\ 
    KD \citep{hinton2015distilling}           &$\underline{91.63}$  &$86.56$&$\underline{83.35}$& $\underline{64.98}$&$81.12$ & 81.52\\

    \textit{Projector} \citep{jiao-etal-2020-tinybert}& $90.88$ & $86.66$& \boldmath$83.73$& $64.14$&$81.98$& 81.47 {\color{red} \small (-0.05)}\\   
    \textit{CKA} \citep{dasguptaimproving} & 91.32 & $\underline{86.93}$ 
 & $82.40$ & $64.62$ &\boldmath$82.52$ & 
 \underline{81.55} {\color{green!60!black} \small (+0.03)}
  \\ 
\textbf{\textit{Flex-KD}}     & \textbf{\boldmath$92.67$} 
  & \textbf{\boldmath$87.14$} 
& $83.20$ & $64.86$ &$\underline{82.30}$ 
   &\textbf{82.03} 
   {\color{green!60!black} \small (+0.51)}\\                
\bottomrule
\end{tabular}
\label{table:two}
\end{table*}

\vspace{-1 em}
\paragraph{IMDB.}
Table~\ref{table:one} shows that Flex-KD achieves the best accuracy across all evaluated student architectures, reaching $90.60\%$ with TinyBERT. This corresponds to improvements of up to $0.44\%$ and $1.21\%$ over the strongest feature-distillation baselines, i.e., CKA and Projector, respectively. Flex-KD also exhibits lower variance across seeds, indicating a more stable transfer signal than projector-based alignment, which can be brittle under representation mismatch and occasionally underperform the KD baseline.

\vspace{-1em}
\paragraph{GLUE.}
Results on GLUE are reported in Table~\ref{table:two}. Flex-KD attains the highest average score ($82.03$) across five tasks and outperforms CKA and Projector on four of them, supporting the benefit of preserving task-relevant functional geometry under representation mismatch.

\begin{figure*}[ht]
\centering

\begin{subfigure}[b]{0.23\textwidth}
    \includegraphics[width=\textwidth]{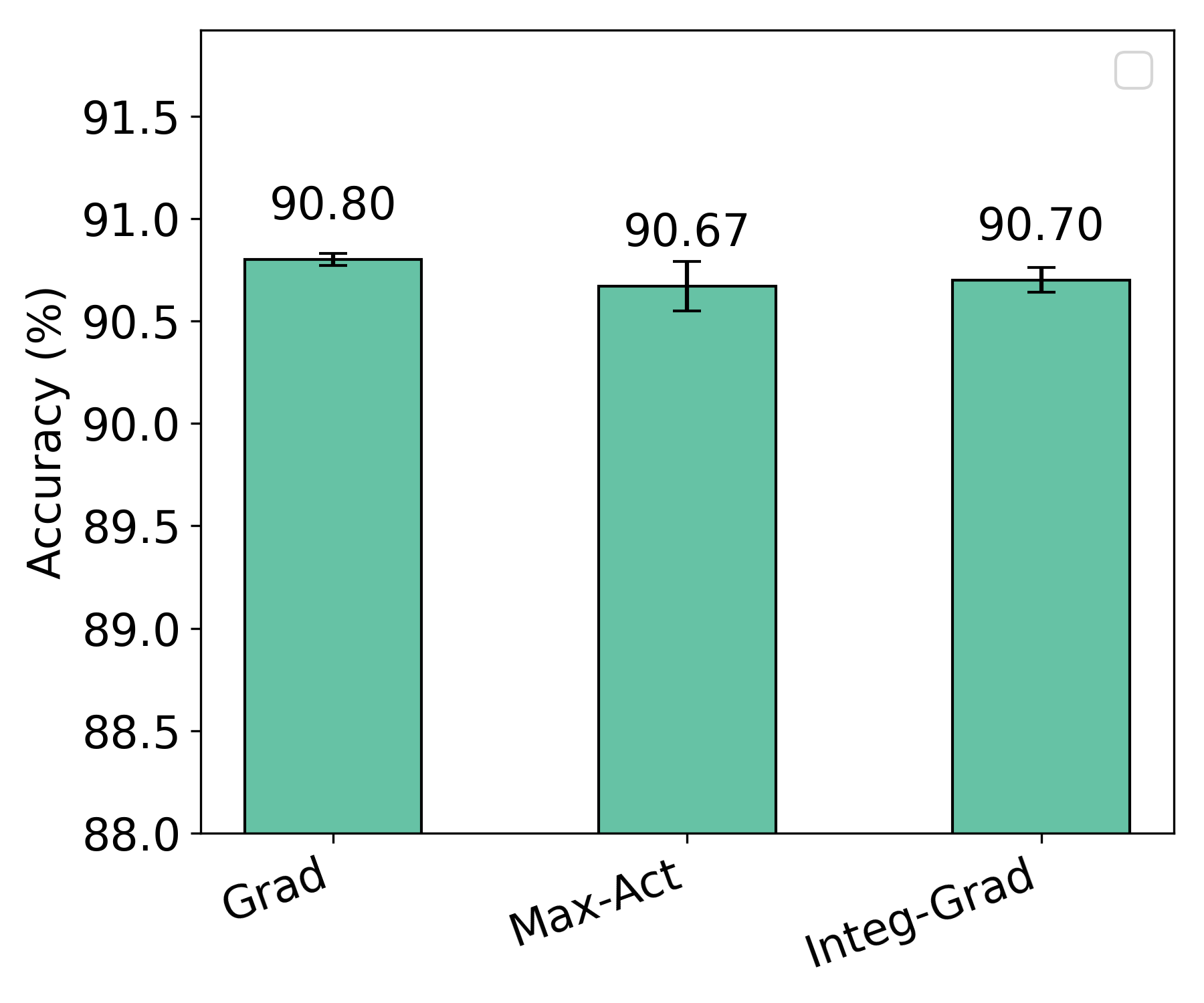}
    \caption{Flex-KD with different unit selection methods on IMDB.}
    \label{fig:threea}
\end{subfigure}
\hfill
\begin{subfigure}[b]{0.49\textwidth}
    \centering
    \begin{subfigure}[b]{0.49\textwidth}
        \includegraphics[width=\textwidth]{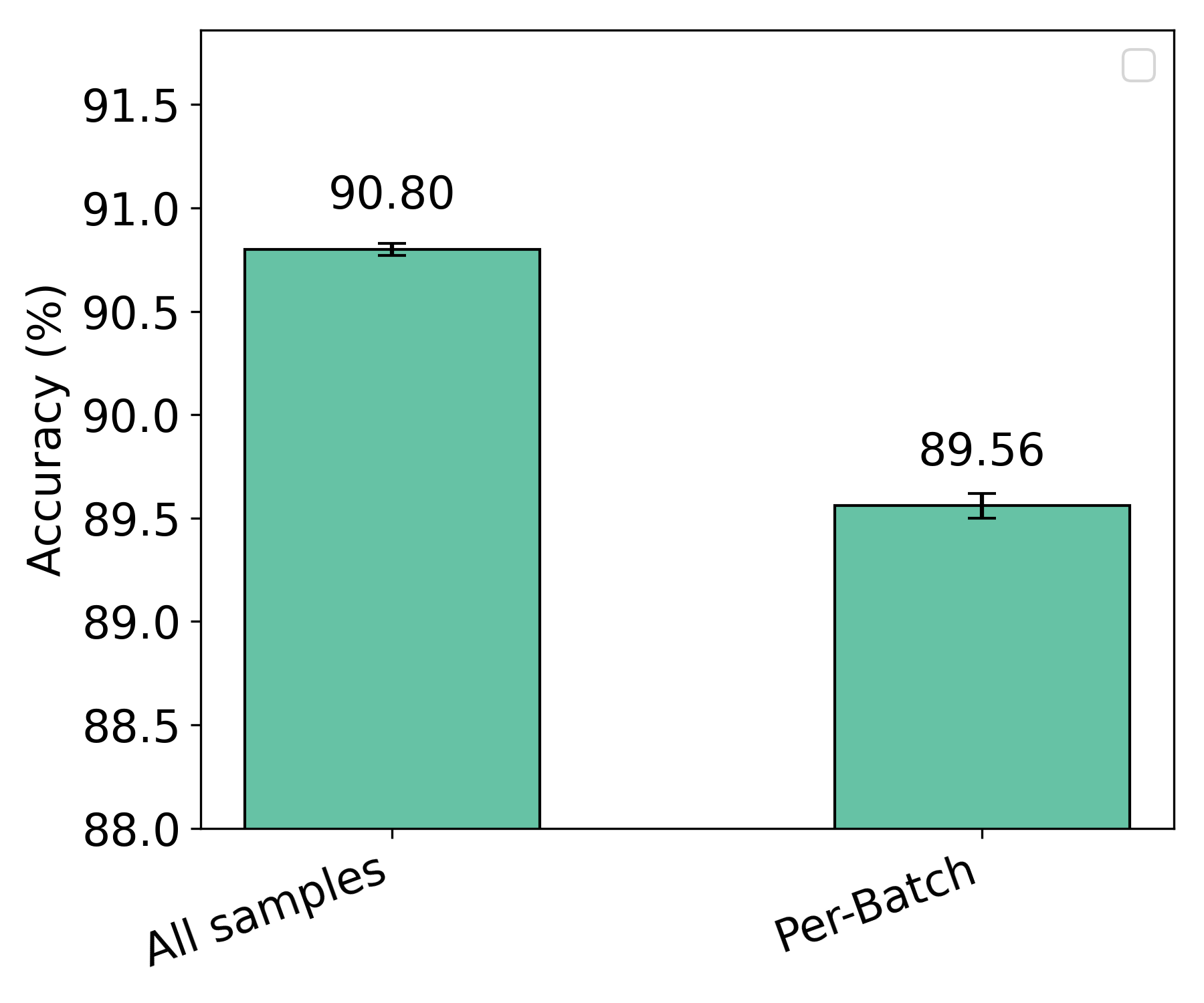}
    \end{subfigure}
    \hfill
    \begin{subfigure}[b]{0.49\textwidth}
        \includegraphics[width=\textwidth]{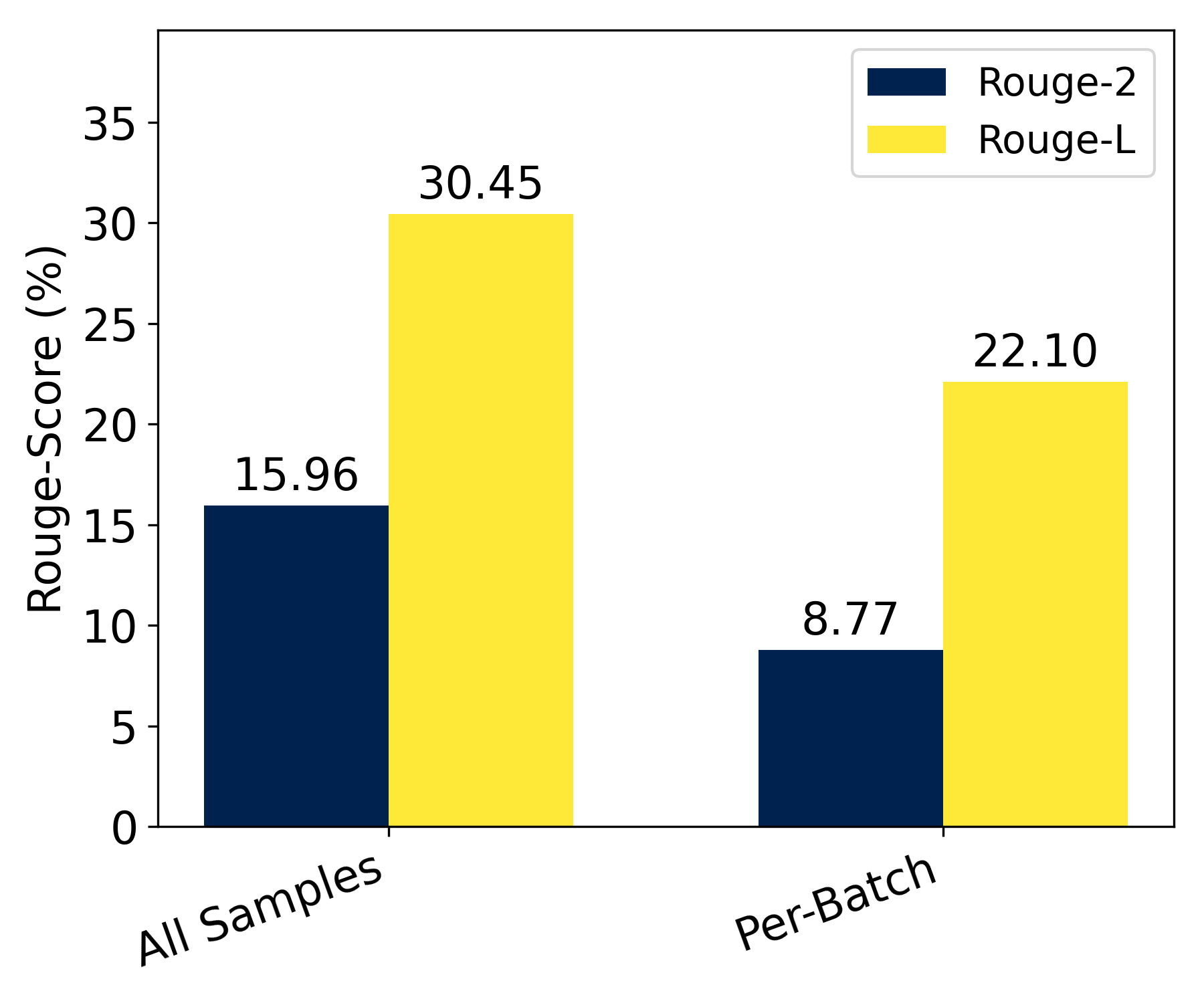}
    \end{subfigure}
    \caption{Flex-KD aggregation methods on IMDB (\textbf{left}) and XSum (\textbf{right}).}
    \label{fig:threeb}
\end{subfigure}
\hfill
\begin{subfigure}[b]{0.27\textwidth}
    \includegraphics[width=\textwidth]{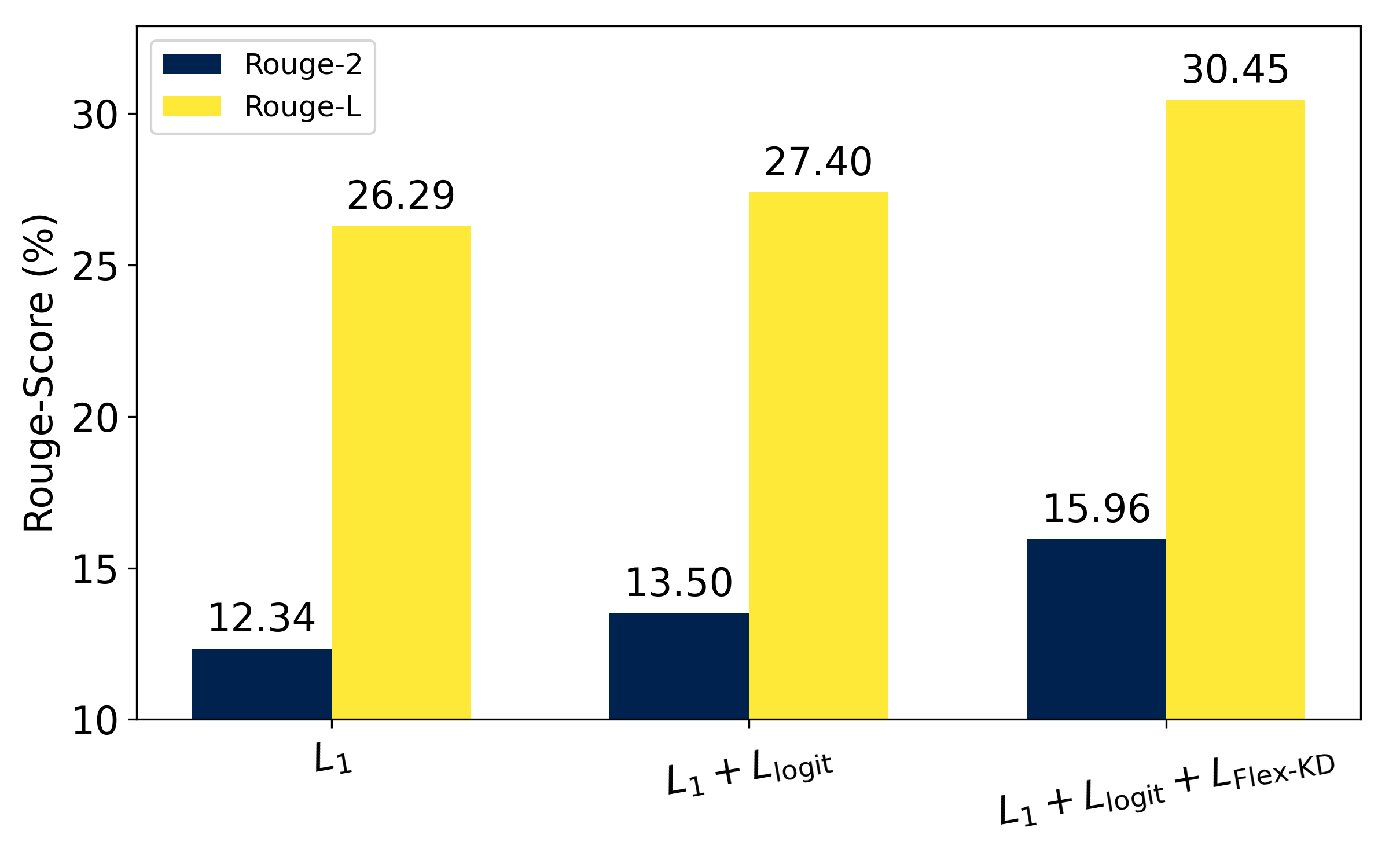}
    \caption{Ablation on loss components (XSum).}
    \label{fig:foura}
\end{subfigure}

\end{figure*}

\textbf{Unit selection strategies.}
In Section~\ref{sec:theory}, we characterize the teacher’s functional geometry through output sensitivity to intermediate representations, which defines a task-dependent local tangent space. Estimating this geometry requires approximating the contribution of individual representation dimensions to the output. We therefore compare several commonly used importance estimators, including activation-based criteria \citep{muralidharan2024compact,zhang2023importance}, standard gradients \citep{iuradaefficient,guo2025comprehensive,song-etal-2024-large}, and integrated gradients \citep{dai2021knowledge}. As shown in Figure~\ref{fig:threea}, all methods achieve similar average performance, but standard gradients consistently yield the best results with the lowest variance. This is consistent with our formulation in Section~\ref{sec:theory}, as gradients directly capture first-order output sensitivity and thus provide the closest approximation of the local functional geometry.

\textbf{Global vs. batch-wise aggregation.}
The functional geometry described in Section~\ref{sec:theory} is defined at the task level rather than for individual samples. Accordingly, we examine whether unit importance should be estimated globally across the training set or locally at the batch level. As shown in Figure~\ref{fig:threeb}, global aggregation consistently outperforms batch-wise selection, with improvements of up to 8\%. We hypothesize that batch-wise aggregation estimates rapidly varying local tangent spaces, introducing non-stationarity into the distillation objective. In contrast, global aggregation provides a more stable approximation of the dominant task-level functional subspace, enabling the student to preserve a consistent geometric structure throughout training.

\textbf{Sensitivity Analysis.} In Figure~\ref{imdbs} of Appendix~\ref{imdbsen}, We investigate the impact of the hyperparameter $\alpha$, which controls the weight of the $L_{\text{Flex-KD}}$ loss, on the student model performance. Flex-KD consistently outperforms the Projector baseline across all $\alpha$ values, demonstrating its robustness.

\vspace{-1 em}
\subsection{Instruction-Following} \label{subsecIS}

We next evaluate Flex-KD on instruction-following tasks~\citep{ouyang2022training}. This setting provides a stringent test of the functional geometry perspective introduced in Section~\ref{sec:theory}, as successful transfer requires preserving how internal representations condition output behavior across diverse prompts, particularly under teacher–student architectural mismatch. For each model family, a teacher is fine-tuned on an instruction–response dataset $D$, and various distillation methods are then applied to train the student on the same task. Full experimental details and baseline descriptions are provided in Appendix~\ref{IF}.

\begin{table*}[!t]

    \centering
    \caption{The Rouge-L score (\%) of the different approaches. * means the result is reported from \citet{gu2024minillm}.
    \#Pars represents the number of parameters. “AVG” represents the average performance across all evaluated tasks. Values highlighted in \textcolor{green!60!black}{green} denote positive performance gains relative to the KD baseline, whereas values in \textcolor{red}{red} indicate negative changes.}
\resizebox{0.83\linewidth}{!}{%
    \begin{tabular}{@{}lllcccccccl@{}}

      \midrule

      Model  &  \#Pars & Method & Dolly & SelfInst& Vicuna &S-NI &UnNI& AVG\\ \midrule

     & 7B  & Teacher & 28.85&20.89	 &18.88&	32.88&36.48	&27.60\\
       \cmidrule(r){2-9}
          &       & FT* \citep{devlin2019bert}                      & 25.85 & 14.59 &17.41&24.13&28.22&22.04\\
             &    & KD* \citep{hinton2015distilling}                   & \textbf{26.17} & 15.13&17.34&24.97&29.22 &22.57\\ 
    Llama    &  1.3B & SeqKD* \citep{kim-rush-2016-sequence}             & \underline{25.98} & 15.00&17.66&25.36&29.83&22.76\\ 

        &         & MiniLLM \citep{gu2024minillm}               & 25.66 & 15.01 &18.34&27.89&32.39&23.86 \\

         &    & \textit{Projector} \citep{jiao-etal-2020-tinybert}    & \textbf{26.17}  &\underline{17.15}&\textbf{19.12}&\underline{30.59}&\underline{34.19}&\underline{25.44} {\color{green!60!black} \small (+1.58)} \\
  &  &\textit{CKA} \citep{dasguptaimproving}   &25.63 
     & 15.83  &18.20
     &28.34  
   &32.87 
    &24.17 
 {\color{green!60!black} \small (+0.31)}\\
 &   & \textbf{\textit{Flex-KD}}      & \underline{25.92} 
   & \textbf{17.21}  &\underline{18.91}  &\textbf{31.23} 
    &\textbf{35.58} &\textbf{25.77} 
    {\color{green!60!black} \small (+1.91)} \\

         \midrule   
       &1.5B & Teacher & 27.20&13.55	 &17.02&	27.46&	32.39&22.92\\
       \cmidrule(r){2-9}
            &     & FT* \citep{devlin2019bert}                      & 23.30 & 10.00&14.70&16.30&18.50&16.56\\ 
           &      & KD* \citep{hinton2015distilling}                  & 22.80 & 10.08&13.40&19.70&22.00 &17.59\\ 
        GPT2 & 120M & SeqKD* \citep{kim-rush-2016-sequence}           & 22.70 & 10.10&14.30&16.40&18.80&16.46\\ 
          &       & MiniLLM \citep{gu2024minillm}                & \underline{24.18} & \textbf{12.33} &\textbf{17.92}&22.67&24.60&\underline{20.34} \\

&  & \textit{Projector} \citep{jiao-etal-2020-tinybert}    &23.60   &11.36&17.61&21.78&\underline{24.63}&19.79 {\color{red} \small (-0.55)}\\
  & & \textit{CKA} \citep{guo2025comprehensive}     & 24.07 &  12.15 &\underline{17.83} 
 &\underline{22.74} 
  &24.35 &20.22 {\color{red} \small (-0.12)}\\
& & \textbf{\textit{Flex-KD}}      &\textbf{24.45} 
  & \underline{12.30} 
  &17.62 &\textbf{23.01} 
   & \textbf{24.87} 
  &\textbf{20.45} 
   {\color{green!60!black} \small (+0.11)}\\
 \midrule

     & 6.7B  & Teacher &28.48 &16.74	 &18.23& 29.92	&32.64&25.20\\
       \cmidrule(r){2-9}
           &      & FT* \citep{devlin2019bert}                     &  \textbf{26.00}& 11.40&15.60&23.10&\underline{28.40}&20.90\\ 

        OPT     &    & MiniLLM \citep{gu2024minillm}                 & 25.50 & \underline{13.54} &\textbf{17.47}&\underline{24.57}& 27.46& \underline{21.70}\\ 

       &       1.3B & \textit{Projector} \citep{jiao-etal-2020-tinybert}      & 25.44  &12.77&16.84&23.98&26.62&21.13 {\color{red} \small (-0.57)} \\
 &   & \textit{CKA} \citep{dasguptaimproving}    &25.25 
   & 12.84 &\underline{17.25} 
     &24.36  
   &27.20 
   &21.38 
{\color{red} \small (-0.32)}\\
 &   & \textbf{\textit{Flex-KD}}      & \underline{25.54} 
   & \textbf{14.31}  &17.06  &\textbf{25.77} 
     &\textbf{28.72}  &\textbf{22.28} 
    {\color{green!60!black} \small (+0.58)} \\

        \bottomrule
        \label{tab:three}
    \end{tabular}}
\vspace{-2.0 em}  
\end{table*}

As shown in Table~\ref{tab:three}, Flex-KD outperforms feature-distillation baselines across datasets and model sizes. For GPT2-small, it achieves the best average ROUGE-L ($20.45\%$), beating Projector on all datasets and CKA on four of five datasets. The same pattern holds for OPT-1.3B, where Flex-KD attains the best average performance and exceeds Projector by up to $+2.10\%$ (UnNI) and $+1.79\%$ (S-NI), with gains over CKA reaching $+1.52\%$. On LLaMA, Flex-KD ranks first on three datasets and second on the remaining two, outperforming Projector and CKA by up to $+2.71\%$ (UnNI) and $+2.89\%$ (S-NI).

Overall, while representation-similarity and projector-based methods often fail to reliably improve instruction-following performance, Flex-KD consistently provides positive gains, supporting the importance of preserving task-relevant functional geometry for conditional generation.

\textbf{Compatibility with logit-level distillation.}
We further evaluate Flex-KD in combination with recent logit KD objectives, including GKD \citep{agarwal2024policy} and DistiLLM \citep{ko2024distillm}. Results in Table~\ref{tab:gkd} of Appendix~\ref{IF} show that Flex-KD consistently improves performance across objectives, indicating that functional geometry alignment is complementary to diverse logit-level KD losses.

\vspace{-1em}
\subsection{Summarization}

We next evaluate Flex-KD on text summarization, following the setup of \citet{dasguptaimproving}. This task provides a challenging test of the functional geometry perspective, as effective summarization requires preserving how encoder and decoder representations jointly influence long-form generation, rather than matching representations at individual layers. Such dependencies are particularly sensitive to architectural mismatch and aggressive compression.

 Full experimental details are provided in Appendix~\ref{sum}. As shown in Table~\ref{tab:four}, Flex-KD consistently outperforms all baselines. Under a $5.5\times$ compression ratio, it improves ROUGE-L by up to $3.75$ points over Projector, which degrades performance relative to logit KD on both datasets. Flex-KD also surpasses CKA, with gains up to $0.96$ ROUGE-2 and $1.25$ ROUGE-L on XSum. Notably, although Flex-KD transfers knowledge only between the final encoder and decoder layers, it consistently outperforms multi-layer distillation methods that align representations across all student layers. This suggests that preserving task-relevant functional geometry is more critical than enforcing layer-wise similarity, particularly under substantial model compression. Results with deeper-layer variants are reported in Appendix~\ref{deepl}.
\vspace{-1em}
\renewcommand{\arraystretch}{1.2}
\begin{table}[ht]
\centering
\large
\caption{
\small ROUGE-2 (R2) and ROUGE-L (RL) scores for different BART students on the CNN/DailyMail and XSum datasets. Every BART student has an equal number of encoder and decoder layers. All baseline results are taken from \citet{dasguptaimproving}.
 Values in \textcolor{green!60!black}{green} indicate a positive performance gain over the KD baseline.
}
% \footnotesize
\resizebox{1\linewidth}{!}{%
\begin{tabular}{lccccc}
\toprule
\textbf{Model} & \textbf{R2(CNN)} & \textbf{RL(CNN)} & \textbf{R2(XSum)} & \textbf{RL(XSum)} \\
\midrule
BART-large (24 $\times$ 1024) & 21.00 & 30.60 & 21.80 & 36.50 \\
\midrule
KD (6 $\times$ 640) \citeyearpar{hinton2015distilling} &  15.10 & 25.80 & 13.50 & 27.40 \\
\textit{Projector} (6 $\times$ 640) \citeyearpar{jiao-etal-2020-tinybert}  & 14.80 {\color{red} \small (-0.30)} & 25.60 {\color{red} \small (-0.20)}  & 12.70 {\color{red} \small (-0.80)} & 26.70 {\color{red} \small (-0.70)}  \\

\textit{CKA} (6 $\times$ 640) \citeyearpar{dasguptaimproving} &  \underline{16.80} {\color{green!60!black} \small (+1.70)} & \underline{26.80} {\color{green!60!black} \small (+1.00)} & \underline{15.00} {\color{green!60!black} \small (+1.50)} & \underline{29.20} {\color{green!60!black} \small (+1.80)} \\
\textbf{\textit{Flex-KD}} (6 $\times$ 640) & \textbf{17.38} {\color{green!60!black} \small (+2.28)} &\textbf{  27.62} {\color{green!60!black} \small (+1.82)} & \textbf{15.96} {\color{green!60!black} \small (+2.46)}& \textbf{30.45} {\color{green!60!black} \small (+3.05)}\\
\midrule
KD (6 $\times$ 768) \citeyearpar{hinton2015distilling}  & 16.40 & 26.80 & 15.10 & 29.20 \\
\textit{Projector} (6 $\times$ 768) \citeyearpar{jiao-etal-2020-tinybert}   & 15.50  {\color{red} \small (-0.90)} & 26.20 {\color{red} \small (-0.60)} & 14.10 {\color{red} \small (-1.00)} & 28.20 {\color{red} \small (-1.00)} \\

\textit{CKA} (6 $\times$ 768) \citeyearpar{dasguptaimproving} & \textbf{17.70} {\color{green!60!black} \small (+1.30)} & \underline{27.70} {\color{green!60!black} \small (+0.90)} & \underline{16.50} {\color{green!60!black} \small (+1.40)} & \underline{31.00} {\color{green!60!black} \small (+1.80)} \\
\textbf{\textit{Flex-KD}} (6 $\times$ 768) & \textbf{17.70} {\color{green!60!black} \small (+1.30)} &  \textbf{27.96} {\color{green!60!black} \small (+1.16)}&\textbf{ 16.65} {\color{green!60!black} \small (+1.55)} & \textbf{31.13} {\color{green!60!black} \small (+1.93)}\\
\bottomrule
\end{tabular}}
\vspace{-1.1 em}
\label{tab:four}
\end{table}

\textbf{Ablation study.}
We conduct an ablation study on XSum summarization to assess the contribution of each component of the proposed objective, using BART (6$\times$640) as the student model. As shown in Figure~\ref{fig:foura}, removing any individual term leads to a clear degradation in performance, indicating that the components play complementary roles in effective knowledge transfer. 

\textbf{Multi-Layer distillation.} In Figure~\ref{fig:fourb} of Appendix~\ref{multi-layer}, we evaluate Flex-KD under a multi-layer distillation setup. Our results show that applying Flex-KD solely on the final hidden layer is sufficient to achieve strong performance, leading to the best performance/efficiency balance. 
This contrasts with methods such as CKA~\citep{dasguptaimproving}, which require distillation across all hidden layers.

\textbf{Consistency of selected units.} In Appendix~\ref{overlap}, we evaluate the stability and consistency of the selected units across five random seeds. Figure~\ref{fig:index_distributions} demonstrate that our unit selection strategy is both stable and consistent, even with a limited number of samples.

\textbf{Compute cost.} As shown in Appendix~\ref{compute cost}, Flex-KD introduces only a negligible one-time gradient computation overhead and, during training, is substantially faster than projector-based and full-layer CKA methods. This demonstrates that Flex-KD is efficient both in terms of overhead cost and end-to-end training time.
\vspace{-1em}
\subsection{Robustness to Data Scarcity}
\begin{table}[h]
    \centering
    \caption{The Rouge-L score of the different approaches with 5\% of the dolly data.}
\resizebox{0.95\linewidth}{!}{%
    \begin{tabular}{@{}lccccc@{}}
        \toprule
       Method & Dolly & SelfInst& Vicuna &S-NI &UnNI \\ 
       \midrule
Teacher & 21.84	&12.74&	15.63&	22.87&	25.99 \\ 
       \cmidrule(r){1-6}
Projector \citeyearpar{jiao-etal-2020-tinybert} &      22.13	&10.72&	16.95&	19.80&	20.81 \\ 
CKA \citeyearpar{dasguptaimproving}      & \underline{22.51} & \textbf{10.89}    &\textbf{17.65   } &	\underline{20.37}    &\underline{21.98}  \\ 
\textbf{Flex-KD}   &   \textbf{23.07}   &	\underline{10.84} &	\underline{17.09} &	\textbf{21.50	} &\textbf{22.58}  \\

        \bottomrule
    \end{tabular}}
    \label{tab6}
    \vspace{-1.5 em}
\end{table}
Flex-KD is motivated by the observation that a model’s output variation is often dominated by a small subset of representation dimensions. As formalized in Equation~\eqref{eq:tail_bound}, the expected deviation in the teacher’s output can be decomposed into contributions from the dominant coordinate set $E_k$ and a residual tail term. Under this formulation, effective distillation does not require matching representations exhaustively, but rather identifying and aligning the coordinates that contribute most strongly to the output.

In low-resource settings, distillation is performed using fewer training examples, resulting in fewer functional matching constraints. A natural question is therefore whether the dominant coordinate set $E_k$ can still be identified reliably from limited data, and whether aligning the student within this subspace suffices to preserve the teacher’s behavior. 
To study this regime, we perform distillation using only 5\% of the Dolly dataset. We follow the same GPT2-based instruction-following setup as in Section~\ref{subsecIS}. Despite the substantially reduced number of samples, Flex-KD consistently outperforms competing feature-distillation approaches. As shown in Table~\ref{tab6}, Flex-KD achieves an average ROUGE-L score of $19.01\%$, compared to $18.08\%$ and $18.68\%$ for the baselines, and matches or surpasses them on most individual tasks. These results suggest that the dominant functional coordinates remain sufficiently stable to be identified even from limited data, enabling effective knowledge transfer in low-resource regimes.

\vspace{-0.90 em}

\section{Conclusion} 
In this work, we revisited feature-based KD from a functional viewpoint, arguing that intermediate representations are only meaningful when they influence the model’s outputs. Building on this perspective, we characterized distillation as preserving a teacher’s functional geometry, the dominant directions in representation space that locally drive output variation, and introduced the notion of an effective functional dimension that is task dependent. Guided by this principle, we proposed Flex-KD, a simple and parameter-free method that (i) estimates task-relevant directions via gradient-based functional contributions, (ii) selects a subspace matched to the student’s representational capacity, and (iii) transfers geometry through correlation-based alignment without learned projectors. Across language understanding and generation benchmarks, Flex-KD consistently improves over strong Feature distillation baselines under severe teacher–student dimension mismatch and in low-resource regimes.
\newpage

\section*{Impact Statement}
This paper presents work whose goal is to advance the field of Machine
Learning. There are many potential societal consequences of our work, none
which we feel must be specifically highlighted here.

\nocite{langley00}

\bibliography{example_paper}
\bibliographystyle{icml2026}

%%%%%%%%%%%%%%%%%%%%%%%%%%%%%%%%%%%%%%%%%%%%%%%%%%%%%%%%%%%%%%%%%%%%%%%%%%%%%%%
%%%%%%%%%%%%%%%%%%%%%%%%%%%%%%%%%%%%%%%%%%%%%%%%%%%%%%%%%%%%%%%%%%%%%%%%%%%%%%%
% APPENDIX
%%%%%%%%%%%%%%%%%%%%%%%%%%%%%%%%%%%%%%%%%%%%%%%%%%%%%%%%%%%%%%%%%%%%%%%%%%%%%%%
%%%%%%%%%%%%%%%%%%%%%%%%%%%%%%%%%%%%%%%%%%%%%%%%%%%%%%%%%%%%%%%%%%%%%%%%%%%%%%%
\newpage
\appendix
\onecolumn

\section{Related Work}\label{sec:RL}

\textbf{Knowledge distillation} \citep{schmidhuber1992learning,hinton2015distilling} is a widely used model compression technique that transfers knowledge from a large teacher model to a small, efficient student model \citep{sanh2019distilbert,gou2021knowledge}. In natural language processing (NLP), KD has been predominantly applied to text classification tasks by aligning the student model with the teacher's output distributions \citep{liangmixkd,zhang2023not}, hidden representations \citep{sun-etal-2019-patient,jiao-etal-2020-tinybert}, or attention matrices \citep{wang2020minilm,wang-etal-2021-minilmv2}. These approaches effectively reduce model size while preserving performance, making them suitable for resource-constrained setups. However, the application of KD in language generation tasks is more complex than in classification tasks \citep{gu2024minillm}. Unlike the fixed-label space of classification, open-ended text generation involves producing discrete token sequences of varying lengths, which adds inherent complexity. 

\textbf{Logit distillation} \citep{hinton2015distilling} aims to minimize the distance between student and teacher output distributions. Current KD techniques for generative models are mainly centered around logit-based methods, where they primarily minimize the forward Kullback-Leibler divergence (FKLD) \citep{kullback1951kullback} between the teacher and student model output distributions \citep{sanh2019distilbert, kim2024token}. This may involve supervision using the teacher's outputs at each generation step \citep{kim-rush-2016-sequence, taori2023stanford}, training on teacher-generated text \citep{peng2023instruction}, or employing reverse Kullback-Leibler divergence (RKLD)\citep{gu2024minillm,kim2024promptkd,guminiplm}, which makes the student distribution focus on certain modes in the teacher’s distribution. Recent work \citep{wangabkd,ko2024distillm} has found that the performance difference of FKLD and RKLD closely depends on the dataset and the task at hand.

\textbf{Feature distillation} \citep{muralidharan2024compact, jiao-etal-2020-tinybert} has received less attention in generative tasks \citep{muralidharan2024compact} compared to logit-based methods, which can be explained by the inherent limitation of conventional feature KD approaches that enforce equal hidden dimensionalities between teachers and students. This restriction reduces both student architectural flexibility and compressibility. A common workaround, adapted from vision and classification \citep{chen2022improved,jiao-etal-2020-tinybert}, is to train an additional linear projector to align the teacher’s and student’s feature spaces \citep{jiao-etal-2020-tinybert}. While effective in pre-training \citep{jiao-etal-2020-tinybert}, projectors often under-perform in downstream tasks \citep{dasguptaimproving}, where data is scarce, introduce extra parameters, and may distort teacher features. The work closest to ours is \citet{dasguptaimproving}, which also tackles the problem of feature distillation between teacher and student with differing hidden sizes. Their approach introduces a metric to compute similarity between tensors of mismatched dimensions, enabling flexible hidden state distillation. However, it still uniformly transfers knowledge from all teacher components without considering their task relevance. As shown in Table~\ref{tab:three}, this limitation can degrade student performance in several cases. This motivates the need for filtering out non-relevant units and focusing on the task-relevant subspace, the core idea of our proposed approach.

\section{Proof of Coordinate-wise Functional Approximation}
\label{app:coord_proof}

\subsection{Proof of Equations~\eqref{eq:coord_bound_1}-\eqref{eq:tail_bound}}

\textbf{ Equations~\eqref{eq:coord_bound_1}-\eqref{eq:tail_bound} (restated)}  For any perturbation $\delta\in\mathbb{R}^{d_T}$, the expected deviation in the teacher’s output satisfies
\begin{equation}
\mathbb{E}_x\!\left[
\left| f_T(h^T(x)+\delta) - f_T(h^T(x)) \right|
\right]
\le
\sum_{i=1}^{d_T} G_i\,|\delta_i|
+
\mathcal{O}(\|\delta\|_2^2).
\end{equation}
Moreover, restricting attention to the top-$k$ representation coordinates defined by $E_k$ yields
\begin{align}
\mathbb{E}_x\!\left[
\left| f_T(h^T(x)+\delta) - f_T(h^T(x)) \right|
\right]
 \le
 \sum_{i\in E_k} G_i\,|\delta_i|  
 +
\mathrm{Tail}(k) +
\mathcal{O}(\|\delta\|_2^2).
\end{align}
\begin{proof}
Let $f_T$ be differentiable with respect to the representation $h^T(x)$.
By first-order Taylor expansion around $h^T(x)$,
\begin{equation}
f_T(h^T(x)+\delta)
=
f_T(h^T(x)) + J_T(x)\,\delta + R(x,\delta),
\end{equation}
where the remainder term satisfies
$|R(x,\delta)| \le C\|\delta\|_2^2$ for some constant $C>0$ in a neighborhood of $h^T(x)$.

Taking absolute values yields
\begin{equation}
\left| f_T(h^T(x)+\delta) - f_T(h^T(x)) \right|
\le
|J_T(x)\,\delta| + C\|\delta\|_2^2.
\end{equation}

Using the definition of the Jacobian,
\begin{equation}
|J_T(x)\,\delta|
=
\left| \sum_{i=1}^{d_T}
\frac{\partial f_T(x)}{\partial h^T_i}\,\delta_i
\right|
\le
\sum_{i=1}^{d_T}
\left|
\frac{\partial f_T(x)}{\partial h^T_i}
\right|
\,|\delta_i|.
\end{equation}

Taking expectations over the data distribution gives
\begin{align}
\mathbb{E}_x
\left[
\left| f_T(h^T(x)+\delta) - f_T(h^T(x)) \right|
\right]
&\le
\sum_{i=1}^{d_T}
\mathbb{E}_x
\left|
\frac{\partial f_T(x)}{\partial h^T_i}
\right|
\,|\delta_i|
+ C\|\delta\|_2^2 \\
&=
\sum_{i=1}^{d_T} G_i\,|\delta_i|
+ C\|\delta\|_2^2, \label{eq:coord_bound}
\end{align}
which proves Equation \eqref{eq:coord_bound_1}.
\qed

Let $E_k$ denote the indices of the top-$k$ entries of $G$, and decompose
\[
\sum_{i=1}^{d_T} G_i |\delta_i|
=
\sum_{i\in E_k} G_i |\delta_i|
+
\sum_{i\notin E_k} G_i |\delta_i|.
\]

If perturbations are restricted to the selected coordinates (i.e., $\delta_i=0$ for $i\notin E_k$),
the second term vanishes.
More generally, for bounded perturbations $|\delta_i|\le 1$, the residual contribution is upper bounded by
\[
\sum_{i\notin E_k} G_i |\delta_i|
\le
\sum_{i\notin E_k} G_i
=
\mathrm{Tail}(k).
\]

Substituting into Eq.~\eqref{eq:coord_bound} yields
\begin{equation}
\mathbb{E}_x
\left[
\left| f_T(h^T(x)+\delta) - f_T(h^T(x)) \right|
\right]
\le
\sum_{i\in E_k} G_i |\delta_i|
+
\mathrm{Tail}(k)
+
\mathcal{O}(\|\delta\|_2^2),
\end{equation}
which completes the proof of Equation~\eqref{eq:tail_bound}.
\qed
\end{proof}

\section{Experiments} \label{exp}
\subsection{Experimental Setup}\label{expdetails}
\textbf{On IMDB.} For the IMDB, we use two distinct teacher–student model pairs. In the first setting, we use BERT-base-uncased (110M parameters) \citep{devlin2019bert}, fine-tuned on IMDB, as the teacher and TinyBERT-General-4L-312D (14M parameters) \citep{jiao-etal-2020-tinybert} as the student. In the second setting, GPT2-medium (345M parameters) \citep{openai2023gpt}, fine-tuned on IMDB, serves as the teacher, with GPT2-base (124M parameters) \citep{openai2023gpt} as the student. The test classification accuracy is reported as the evaluation metric. The teacher models are fine-tuned on the IMDB dataset for $3$ epochs with a batch size of $8$ and an Adam optimizer with a learning rate equal to $5e-5$. 
During the distillation process, the student models are trained for $3$ epochs, the batch size is set to $8$, the optimizer is set to Adam with a learning rate of $5e-5$. Each experiment is repeated for 3 times and the average and the standard deviations are reported. The weight of each KD stand-alone loss and the weight of the hard loss are fixed to 0.5 \citep{sanh2019distilbert,jiao-etal-2020-tinybert}, for instance, for Flex-KD, we have $\alpha = 0.5$, $\beta = 0$, and $\lambda = 0.5$.

\textbf{On GLUE.} For the GLUE benchmark, we selected 5 datasets that cover different categories and sizes, small‑size (RTE, STS‑B), medium‑size (MRPC, SST‑2), and large‑size (MNLI), to ensure varied scenarios. MRPC and STS-B for paraphrase and semantic similarity. SST-2 for sentiment classification, and MNLI and RTE for natural language inference.
We report results for CKA \citep{dasguptaimproving}, which is a recently proposed feature distillation method that does not require the student and the teacher to have equal hidden dimensions, linear projection (Projector) \citep{jiao-etal-2020-tinybert}, vanilla KD (KD) \citep{hinton2015distilling}, and our proposed approach (Flex-KD). In this setting, we use  GPT2-medium (345M parameters), fine-tuned on each of the tasks, as the teacher and GPT2-base (124M parameters) as the student. For MRPC, we report the average of F1 score and accuracy; for STS-B, we report the average of Pearson and Spearman correlations. Accuracy is used as the evaluation metric for the remaining tasks. Here, the teacher is trained for $3$ epochs with batch size $8$, and and Adam optimizer with a learning rate of $5e-5$. For the distillation process, the number of epochs, the batch size, the learning rate are set to $3$, $16$, and $5e-5$, respectively. The weight of each KD stand-alone loss and the weight of the hard loss are fixed to 0.5 \citep{sanh2019distilbert,jiao-etal-2020-tinybert}, for instance, for Flex-KD, $\alpha = 0.5$, $\beta = 0$, and $\lambda = 0.5$. For the vanilla-KD (KD), the logit loss weight was set to $0.1$. All experiments are repeated for three random seeds and the average and the standard deviations are reported. For fair comparison, in this task, we used the same correlation loss as for our proposed method for the projector approach. As demonstrated in prior work \citep{chen2022improved, jiao-etal-2020-tinybert}, projecting the student’s feature representations into the teacher’s feature space yields superior performance compared to the reverse direction. Accordingly, in our implementation of the projector-based approach, we align the student’s hidden representations to those of the teacher via a learned projection. 

\subsection{Loss} \label{loss}
In Table~\ref{table:ten}, we conduct an experiment comparing the performance of three feature distillation loss functions: MSE, cosine distance, and a correlation-based loss on the IMDB classification task. The teacher model is BERT-base (110M) and the student model is TinyBERT (14M). As shown in the table, the student model trained with the correlation-based loss achieves better performance and exhibits lower standard deviation, demonstrating its effectiveness.

\begin{table*}[h!]
\centering

\caption{Comparison between MSE, Cosine distance and Correlation as feature loss functions.
} 
\begin{tabular}{@{}lccc@{}}
\toprule
\multirow{2}{*}{Method} &  \multicolumn{3}{c}{110M $\rightarrow$ 14M (BERT)} \\
                        & MSE& Cosine & Correlation \\
\midrule
Teacher                &  \multicolumn{3}{c}{94.14}         \\
\cmidrule(r){1-4}
\textbf{Flex-KD}         & $89.94\pm0.07$ & $90.05\pm0.13$& \boldmath$90.80\pm0.03$\\
\bottomrule
\end{tabular}
\label{table:ten}
\end{table*}

\subsection{Performance on GLUE} \label{perglue}
In Table~\ref{table:six}, we present the performance of different KD approaches on several tasks from the glue benchmark. Results (in \%) are averaged over three random seeds. The teacher model is GPT2-medium (345M parameters), and the student model is GPT2-small (124M parameters). “AVG” represents the average performance across all evaluated tasks. For feature distillation, here we distill from a hidden size of 1024 to a hidden size of 768.

\begin{table*}[h!]
    \centering
    \caption{Results (\%) are averaged 
    over 3 random seeds. M for million. Teacher is gpt2-medium and Student is gpt2-small. AVG is for the average performance across all the tasks.}
    \resizebox{1\linewidth}{!}{%
    \begin{tabular}{@{}lcccccc@{}}
        \toprule
 Method & SST-2 & STS-B & MRPC & RTE&MNLI&AVG\\ \midrule   
    Teacher &94.49 &	88.23&  84.09& 68.23&85.10   & 84.02\\
        \cmidrule(r){1-7}

FT                    & $91.32\pm0.29$ &  $86.58\pm0.33$& $81.68\pm1.34$&\boldmath$65.95\pm2.45$ & $81.78\pm0.11$&81.46\\ 
    KD                      &$\underline{91.63\pm0.11}$  &$86.56\pm0.29$&$\underline{83.35\pm0.87}$& $\underline{64.98\pm0.00}$&$81.12\pm0.06$ & 81.52\\ 
    Projector & $90.88\pm0.72$ & $86.66\pm1.45$& \boldmath$83.73\pm0.54$& $64.14\pm1.45$&$81.98\pm0.27$& 81.47\\   
    CKA & 91.32$\pm0.77$ & $\underline{86.93\pm0.03}$& $82.40\pm1.43$& $64.62\pm0.00$&\boldmath$82.52\pm0.27$& \underline{81.55}\\ 
   
\textbf{Flex-KD}      & \textbf{\boldmath$92.67\pm0.13$} & \textbf{\boldmath$87.14\pm0.21$}& $83.20\pm1.19$& $64.86\pm1.98$&$\underline{82.30\pm0.21}$&\textbf{82.03}\\

        \bottomrule
        \end{tabular}}
    \label{table:six}
\end{table*}

\subsection{Sensitivity analysis} \label{imdbsen}
\begin{figure}
  \centering
  \includegraphics[width=0.38\textwidth]{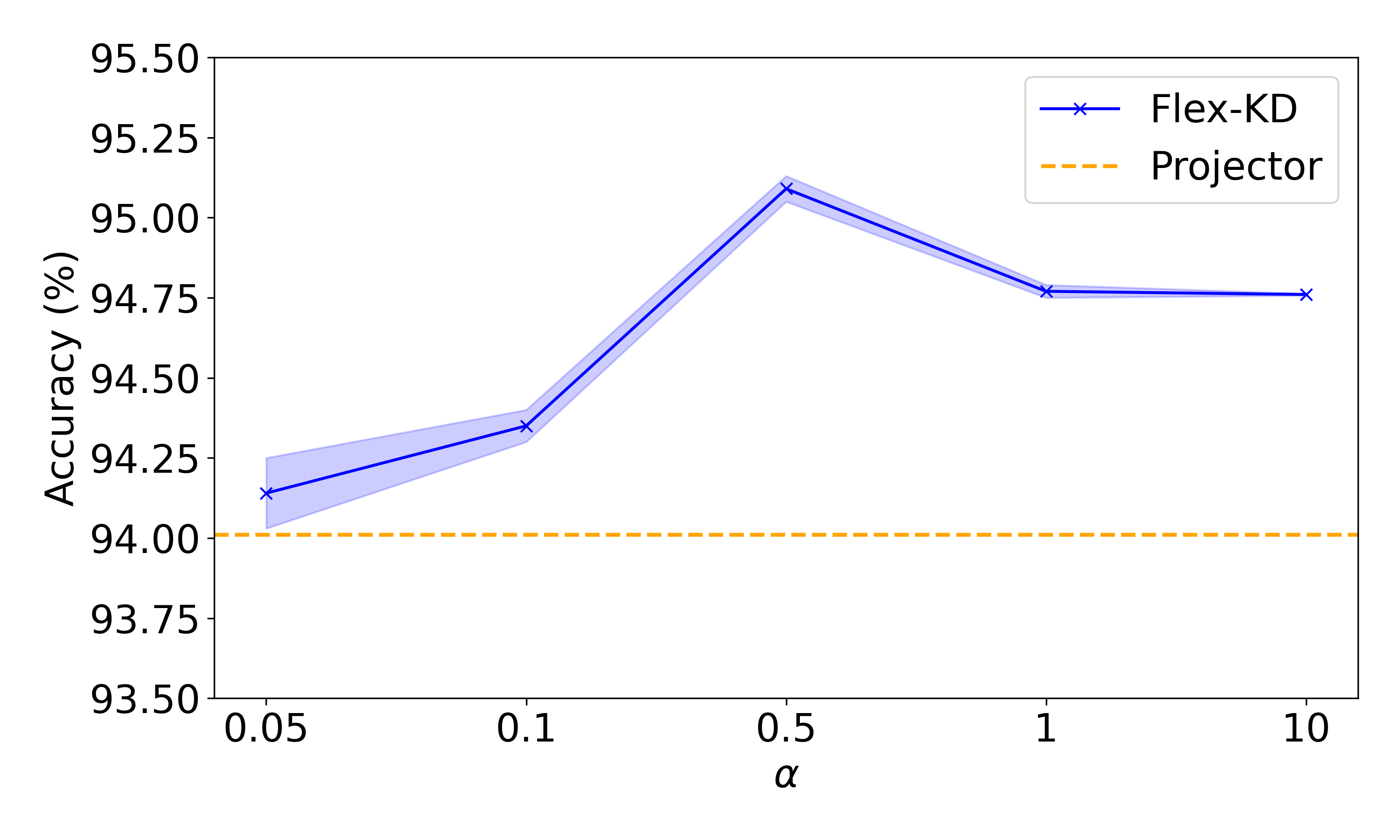}
  % \vspace{-1 em}
  \caption{Student model performance on the IMDB dataset as a function of $\alpha$.}
  \label{imdbs}
  % \vspace{-1em}
\end{figure}

We investigate the impact of the hyperparameter $\alpha$, which controls the weight of the $L_{\text{Flex-KD}}$ loss, on the student model performance on the IMDB dataset. The final training objective is a weighted combination of $L_{\text{Flex-KD}}$ and the supervised cross-entropy loss, where the weight of the supervised component is fixed at 0.5, and $\alpha$ is varied across the range $[0.05, 0.1, 0.5, 1, 10]$. We used the same setup outlined in Section~\ref{classifi}. Each experiment is repeated for 3 random seeds and the average is reported. As shown in Figure~\ref{imdbs}, Flex-KD consistently outperforms the Projector baseline across all $\alpha$ values, demonstrating its robustness and effectiveness.

\subsection{Ablation on Teacher Models}

\begin{table*}[h!]
\centering
\caption{Unified ablation study across RTE, STS-B, and SST-2. 
For STS-B, we report the average of Pearson and Spearman. 
All $\Delta$ values are rounded to three decimal places.}
\label{tab:unified_ablation}
\begin{tabular}{@{}llcccccc@{}}
\toprule
\multirow{2}{*}{\textbf{Ablation Type}} & 
\multirow{2}{*}{\textbf{\% Removed}} & 
\multicolumn{2}{c}{\textbf{RTE}} & 
\multicolumn{2}{c}{\textbf{STS-B (Avg)}} & 
\multicolumn{2}{c}{\textbf{SST-2}} \\
\cmidrule(lr){3-4} \cmidrule(lr){5-6} \cmidrule(lr){7-8}
 &  & Acc & $\Delta$ Acc & Avg & $\Delta$ Avg & Acc & $\Delta$ Acc \\
\midrule

% ---------------- High Importance ----------------
\multirow{4}{*}{High Importance}
 & 0\%  & 68.23 & - & 88.23 & - & 94.49 & - \\
 & 10\% & 60.29 & -7.9 & 87.09 & -1.1 & 94.27 & -0.2 \\
 & 20\% & 55.23 & -13.0 & 87.09 & -1.1 & 94.15 & -0.3 \\
 & 50\% & 54.51 & \textbf{\textcolor{red}{-13.7}} & 81.75 & \textbf{\textcolor{red}{-6.5}} & 93.46 & \textbf{\textcolor{red}{-1.0}} \\
\midrule

% ---------------- Low Importance ----------------
\multirow{4}{*}{Low Importance}
 & 0\%  & 68.23 & - & 88.23 & - & 94.49 & - \\
 & 10\% & 68.59 & +0.4 & 88.23 & 0.0 & 94.50 & 0.0 \\
 & 20\% & 68.23 & 0.0 & 88.21 & 0.0 & 94.50 & 0.0 \\
 & 50\% & 68.59 & \textcolor{green!60!black}{+0.4} & 88.23 & 0.0 & 94.15 & \textcolor{red}{-0.3} \\
\midrule

% ---------------- Random ----------------
\multirow{4}{*}{Random}
 & 0\%  & 68.23 & - & 88.23 & - & 94.49 & - \\
 & 10\% & 68.95 & +0.7 & 88.22 & 0.0 & 94.38 & -0.1 \\
 & 20\% & 69.68 & +1.4 & 88.12 & -0.1 & 93.92 & -0.6 \\
 & 50\% & 66.43 & \textcolor{red}{-1.8} & 87.96 & \textcolor{red}{-0.3} & 94.27 & -0.2 \\
\bottomrule
\end{tabular}
\end{table*}

\subsection{Instruction-Following} \label{IF}

The conducted experiments follow a similar setup to the one outlined in \citet{gu2024minillm}. We evaluate our method on three teacher–student model pairs. First, the Llama model (7B parameters), fine-tuned on the instruction-following Dolly dataset~\footnote{\url{https://github.com/databrickslabs/dolly/tree/master}}, serves as the teacher, with Llama (1.3B parameters) as the student . In the second setting, the GPT2-XL model (1.5B parameters), fine-tuned on the instruction-following Dolly dataset, serves as the teacher, with GPT2-small (124M parameters) as the student. In the third setting, the OPT-6.7B model, fine-tuned on the Dolly dataset, is distilled into the smaller OPT-1.3B student model. We compare our approach against several approaches, including the standard fine-tuning (FT) of the student model, KD \citep{sanh2019distilbert}, SeqKD \citep{taori2023stanford}, MiniLLM \citep{gu2024minillm}, as well as the direct competitive feature KD methods, i.e., Projector \citep{jiao-etal-2020-tinybert} and CKA \citep{dasguptaimproving}. For evaluation metrics, similar to \citet{gu2024minillm}, we report the Rouge-L \citep{lin2004rouge} score on the following benchmark datasets: Dolly test set, SelfInst \citep{wang2022self}, Vicuna \citep{chiang2023vicuna}, S-NI \citep{wang-etal-2022-super}, and UnNI \citep{honovich-etal-2023-unnatural} datasets. The Rouge-L score measures the precision of the model generation and it was shown by \citet{wang-etal-2022-super} that it is suitable for large-scale instruction-following evaluation.

Across all settings, we adopt a consistent distillation framework. The student model is first fine-tuned for 3 epochs, and the checkpoint with the lowest validation loss is used as the initialization point for subsequent distillation. The distillation process is run for 5,000 iterations with a total batch size of 8, using the Adam optimizer configured with an $\epsilon = 1\text{e-}8$, and a weight decay of $1\text{e-}6$. The learning rate is set to $5\text{e-}6$. All reported results are averaged over three random seeds (10, 20, 30) for training and five seeds (10, 20, 30, 40, 50) for evaluation. Except for Llama we only did the evaluation across 5 seeds, i.e.,(10, 20, 30, 40, 50). The final evaluation is always conducted using the last saved checkpoint.

For the MiniLLM baseline, we employ only the reverse Kullback-Leibler divergence distillation loss, as outlined in \citet{gu2024minillm},  between the teacher and student logits as the training objective for the student model. For our method (Flex-KD), as well as the CKA and Projector baselines, we use a combination of the logit distillation loss used to train MiniLLM (RKLD) and the corresponding feature-level distillation loss for the student training. For Feature distillation methods, we distill only the teacher last hidden layer to the student last hidden layer. 

Specifically, for Flex-KD, $\alpha =  0.05$. Following the configurations from \citet{dasguptaimproving}, both CKA and projector baselines use a feature loss weighted by 1. The logit-level loss (reverse KL divergence) is consistently weighted by 1 across all methods. In the projector setup, we used mean squared error (MSE) as the loss function, in line with \citet{dasguptaimproving,jiao-etal-2020-tinybert}.
For teacher model fine-tuning, all teachers are trained for 10 epochs. GPT2 uses a batch size of 8 and a learning rate of $1\text{e-}4$, while OPT and Llama are trained with a total batch size of 8 and a learning rate of $1\text{e-}5$.

The following are some details related to the competitive methods:
\begin{itemize}
    \item \textbf{FT} \citep{devlin2019bert} refers to standard fine-tuning.
    \item \textbf{KD} \citep{sanh2019distilbert} namely, word-level KD, where the student model is trained on the teacher model's output at each token step.
    \item \textbf{SeqKD} \citep{taori2023stanford} refers to sequence-level knowledge distillation, where the student model is trained on data generated by the teacher model.
    \item \textbf{MiniLLM} \citep{gu2024minillm} employs reverse KL divergence to distill knowledge from the teacher model's logits.
\end{itemize}
We evaluate our models on the following instruction-following datasets:
\begin{itemize}
    \item \textbf{Dolly}: 500 samples from the \texttt{databricks-dolly-15K} dataset used as test set.
    \item \textbf{SelfInst} \citep{wang2022self}: A user-oriented instruction-following set consisting of 252 samples.
    \item \textbf{Vicuna} \citep{chiang2023vicuna}: 
The set of 80 difficult questions used for the Vicuna evaluation.

 \item \textbf{S-NI} \citep{wang-etal-2022-super}: The SUPER-NATURALINSTRUCTIONS test set comprises 9K samples spanning 119 tasks. Following \citet{gu2024minillm}, we divide it into three subsets based on ground truth response lengths: $[0,5], [6,10],[11,+\infty]$ and we use the $[11,+\infty]$ subset.

    \item \textbf{UnNI} \citep{honovich-etal-2023-unnatural}: 
   The core set of UNNATURALINSTRUCTIONS comprises 60K samples. Following a similar approach to S-NI, we evaluate on a randomly selected subset of 10K examples from the \([11, +\infty]\) range.
\end{itemize}

\begin{table}[h!]
\centering
\caption{Rouge-L results. Values highlighted in \textcolor{green!60!black}{green} denote positive performance gains relative to the logit baseline, whereas values in \textcolor{red}{red} indicate negative changes. Teacher is GPT2-Xlarge (1.5B) and all students are GPT2-base(120M)}
\label{tab:gkd}
\begin{tabular}{@{}lccc@{}}
\toprule
 Method & S-NI & UnNI & AVG \\
\midrule

% \multirow{4}{*}
   Teacher 
      & 27.46 & 32.39 & 29.92\\ 
\cmidrule(l){1-4}

    GKD \citeyear{agarwal2024policy} 
      & 18.88 & 21.41 & 20.14 \\

    \textit{Projector + GKD} \citeyear{jiao-etal-2020-tinybert} 
      &  15.86 & 16.44 &  16.15 {\color{red} \small (-3.99)} \\

 \textit{CKA + GKD} \citeyear{guo2025comprehensive} 
      & 19.52 & 21.62 & 20.57 {\color{green!60!black} \small (+0.43)} \\

     \textbf{\textit{Flex-KD + GKD}} 
      & \textbf{19.91} & \textbf{22.58 }& \textbf{21.24} {\color{green!60!black} \small (+1.10)} \\
\cmidrule(l){1-4}

   Distillm \citeyear{gu2024minillm} 
      & 25.04 & 27.68 & 26.36\\

     \textit{Projector + Distillm} \citeyear{jiao-etal-2020-tinybert} 
      & 21.60 & 23.74 &  22.67 {\color{red} \small (-3.69)} \\

  \textit{CKA + Distillm} \citeyear{guo2025comprehensive} 
      & 25.55 & 27.55  &  26.55 {\color{green!60!black} \small (+0.19)} \\

    \textbf{\textit{Flex-KD + Distillm}} 
      & \textbf{26.52} & \textbf{28.00} & \textbf{27.26} {\color{green!60!black} \small (+0.90)} \\

\bottomrule
\end{tabular}
% \vspace{-1.5em}
\end{table}

\subsection{Summarization} \label{sum}

In this experiment, we follow a similar experimental setup to that outlined in \citet{dasguptaimproving}. We distill BART-large \citep{lewis2019bart} into smaller student models with varying depth (6 and 12 layers) and hidden dimensionality (640 and 768), and evaluate on CNN/DailyMail \citep{hermann2015teaching} and XSum \citep{narayan2018don}.
The student training objective consists of three components: (1) a supervised cross-entropy loss on the target summary, (2) a logit distillation loss, which is the Kullback-Leibler (KL) divergence loss between the teacher and student output distributions, and (3) a feature distillation loss. For the feature distillation, we evaluate three methods: our proposed Flex-KD, CKA \citep{dasguptaimproving}, and a linear projection-based mean squared error (MSE) (Projector) \citep{dasguptaimproving,jiao-etal-2020-tinybert} . Additionally, we report results for standard logit-level KD \citep{sanh2019distilbert} without any feature distillation.

As described in \citet{dasguptaimproving}, CKA and Projector losses are applied between each student layer and uniformly spaced layers from the teacher model. For the Projector variant, hidden states from the student and teacher are aligned via learned linear projections, followed by MSE as in \citet{jiao-etal-2020-tinybert,dasguptaimproving}. For our Flex-KD, $L_{\text{Flex-KD}}$ \eqref{corr_loss} is only applied between the last encoder layers and the last decoder layers of the student and the teacher models. 
% Additional experimental details are provided in Appendix~\ref{sum}. 

For Flex-KD, we utilize only 640 samples (40 batches) to identify the top task-relevant units in the teacher model and the hyperparameters are set as follows: $\alpha = 0.05$, $\beta = 1$, $\lambda = 1$, and the batch size is 16. Following the setup in \citet{dasguptaimproving}, training is performed using the Adam optimizer with a learning rate of $1 \times 10^{-4}$ and a weight decay of $5 \times 10^{-4}$. The maximum input context length is set to 1,024 tokens, and the output summary is constrained to 128 tokens. All experiments are conducted on a single NVIDIA A100 GPU with 80 GB of memory.

\subsubsection{Summarization on XSum: student model with deeper layers } \label{deepl}
To further evaluate Flex-KD robustness with deeper student models, Table~\ref{tab:five} reports results on the XSum dataset, where Flex-KD achieves consistent improvements over the baselines, including up to a 1.43-point gain in ROUGE-L over Projector.

\begin{table}[h!]
\centering
\caption{R2 and RL for deeper BART students on the XSum dataset. 
All baseline results are taken from \citet{dasguptaimproving}. Teacher is BART-Large (24 $\times$ 1024) 
and student is BART (12 $\times$ 768).}
\label{tab:five}
% \vspace{-0.5em}

\begin{tabular}{lcc}
\toprule
\textbf{Model}  & \textbf{R2(XSum)} & \textbf{RL(XSum)} \\
\midrule
BART-large  &  21.80 & 36.50 \\
\midrule
KD \citep{hinton2015distilling}   & 17.60 & 32.00 \\
\textit{Projector} \citep{jiao-etal-2020-tinybert}  &  17.70 {\color{green!60!black} \small (+0.10)}   & 32.10 {\color{green!60!black} \small (+0.10)}  \\
\textit{CKA} \citep{dasguptaimproving} & \underline{18.70} {\color{green!60!black} \small (+1.10)} 
             & \underline{33.50} {\color{green!60!black} \small (+1.50)} \\
\textbf{\textit{Flex-KD}}  
   & \textbf{18.93} {\color{green!60!black} \small (+1.33)} 
   & \textbf{33.53} {\color{green!60!black} \small (+1.53)}\\
\bottomrule
\end{tabular}
\vspace{-1em}
\end{table}
\subsubsection{Multi-Layer distillation} \label{multi-layer}
\begin{figure*}[h!]
  \centering
  % First subfigure
  
  \hfill
  % Second + Third grouped
  \begin{subfigure}[b]{1.0\textwidth} % take double space
    \centering
    \begin{subfigure}[b]{0.48\textwidth}
      \includegraphics[width=\textwidth]{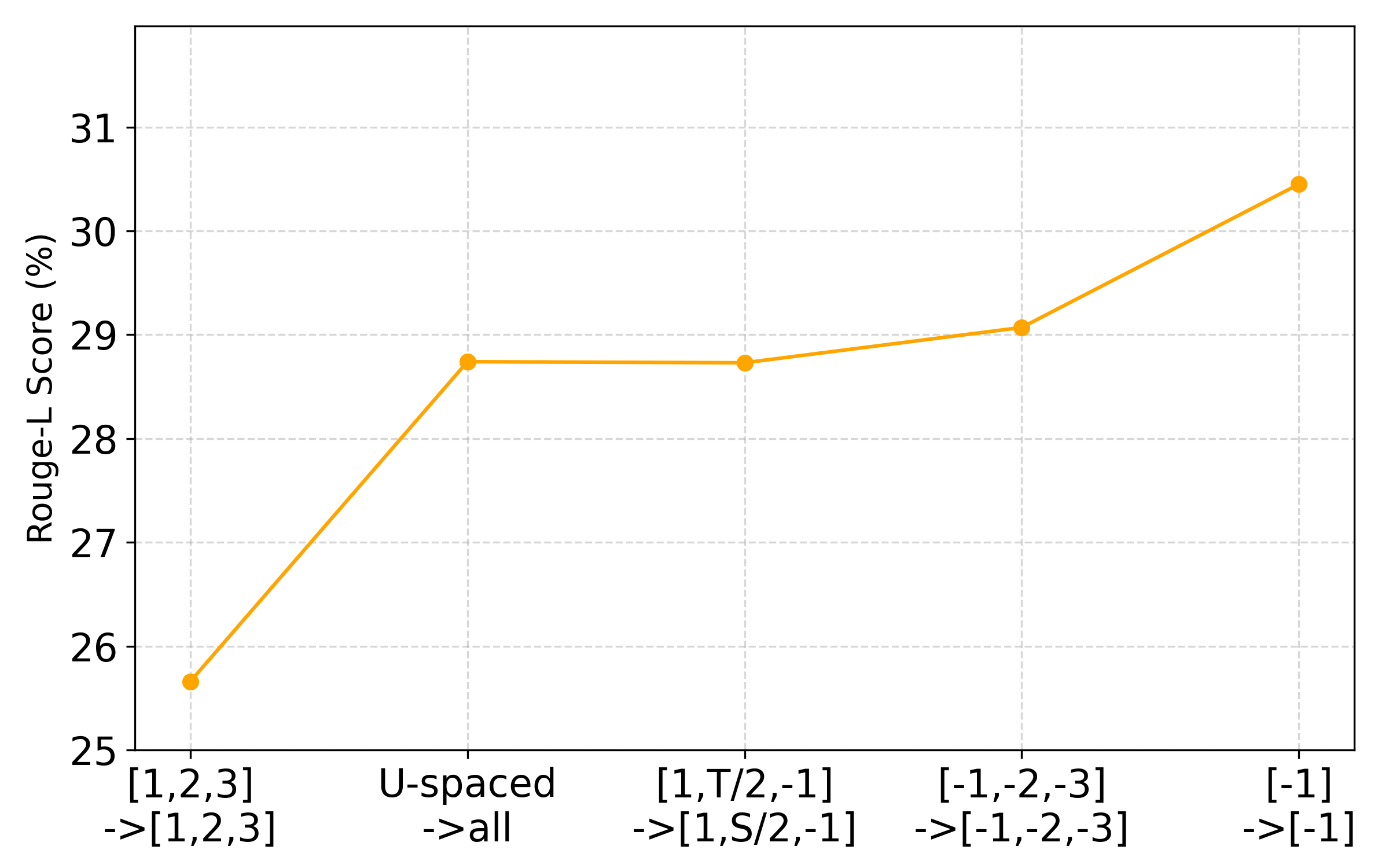}
    \end{subfigure}
    \hfill
    \begin{subfigure}[b]{0.48\textwidth}
      \includegraphics[width=\textwidth]{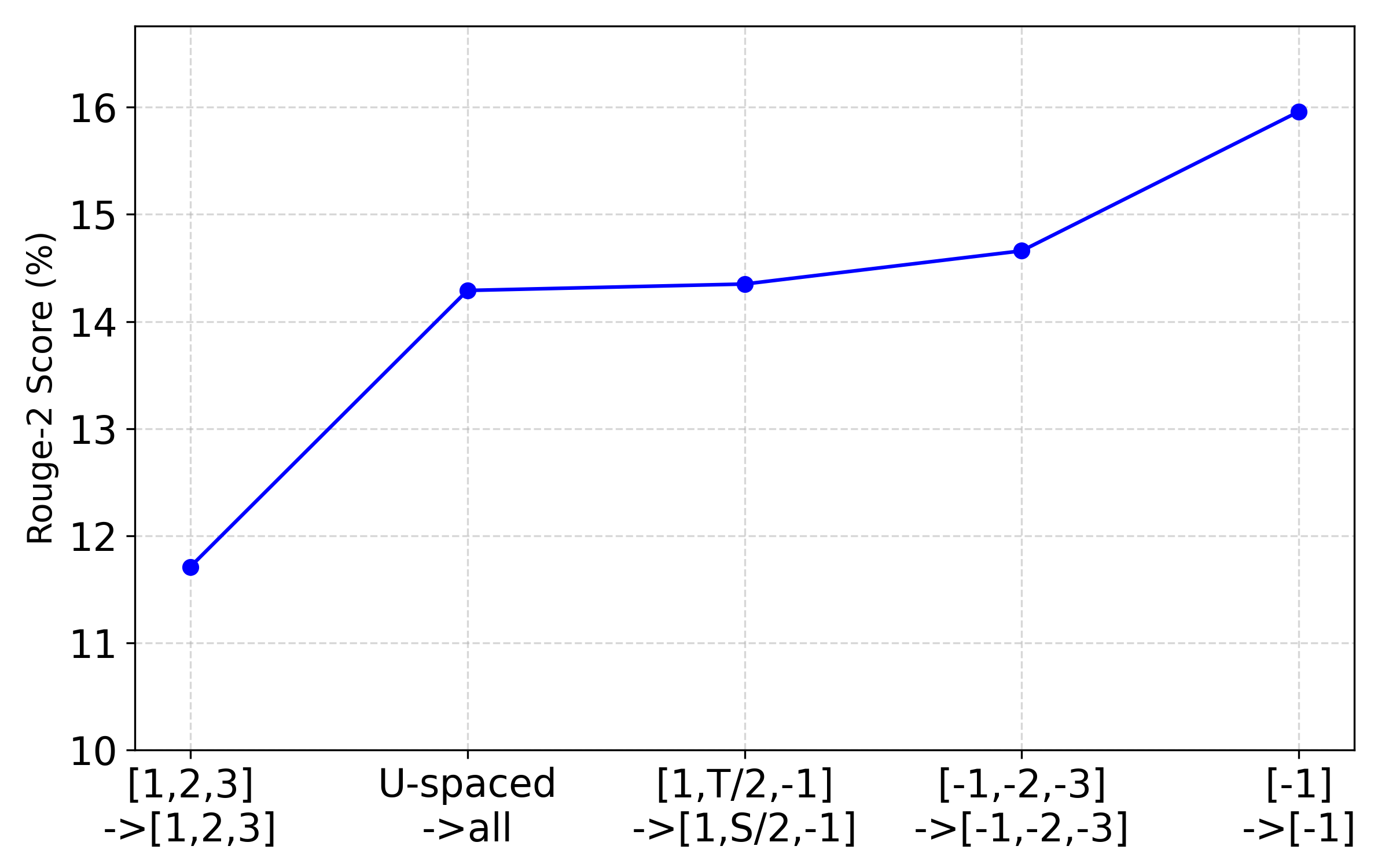}
    \end{subfigure}

        % \vspace{+2em}
  \end{subfigure}

      \caption{Flex-KD with different layers distillation.}
        \label{fig:fourb}
\end{figure*}

In Figure~\ref{fig:fourb}, we evaluate Flex-KD under a multi-layer distillation setup on XSum, where knowledge is transferred from multiple teacher layers to student layers. T and S are the total number of teacher and student hidden layers. The x-axis illustrates the element-wise mapping between teacher and student layers. Our results show that applying Flex-KD solely on the final hidden layer is sufficient to achieve strong performance, leading to the best performance/efficiency balance. This contrasts with methods such as CKA~\citep{dasguptaimproving}, which require distillation across all hidden layers. We argue that multi-layer distillation introduces a more complex optimization process, whereas last-layer distillation is simpler and more efficient. Prior work has also highlighted the importance of the final hidden state in LLMs, showing its significance for both generative and classification tasks \citep{gromov2024unreasonable, men2024shortgpt, saadi2023learn}.

\subsubsection{Summarization on XSum and CNN/DailyMail: Overlap of selected units across random seeds} \label{overlap}
\begin{figure}[h!]
    \centering
    \begin{subfigure}[b]{0.48\textwidth}
        \includegraphics[width=\textwidth]{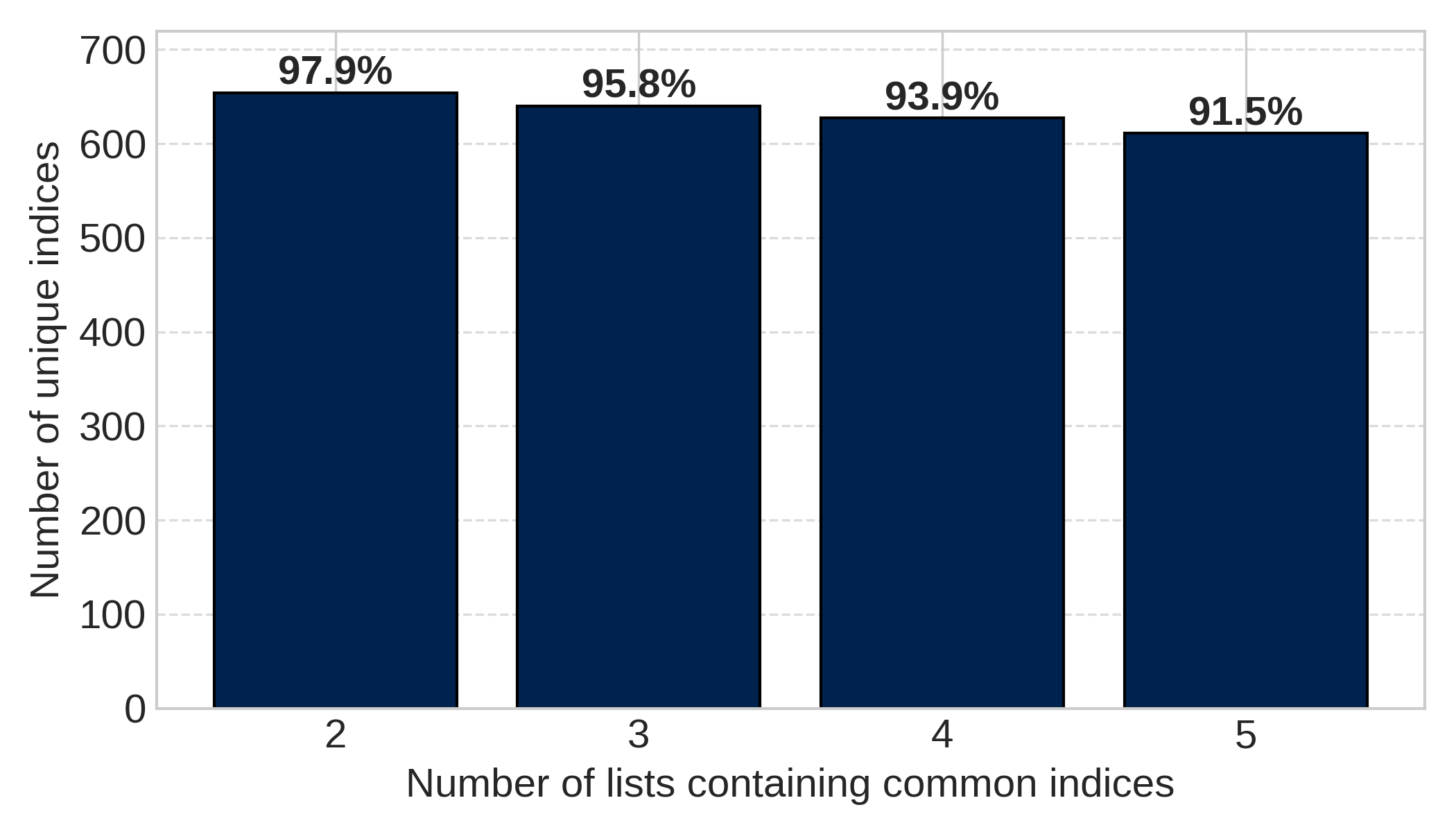}
        \captionsetup{skip=1pt}
        % \caption{XSum distribution}
        \label{fig:xsum}
    \end{subfigure}
    \hspace{0.02\textwidth} % small horizontal space
    \begin{subfigure}[b]{0.48\textwidth}
        \includegraphics[width=\textwidth]{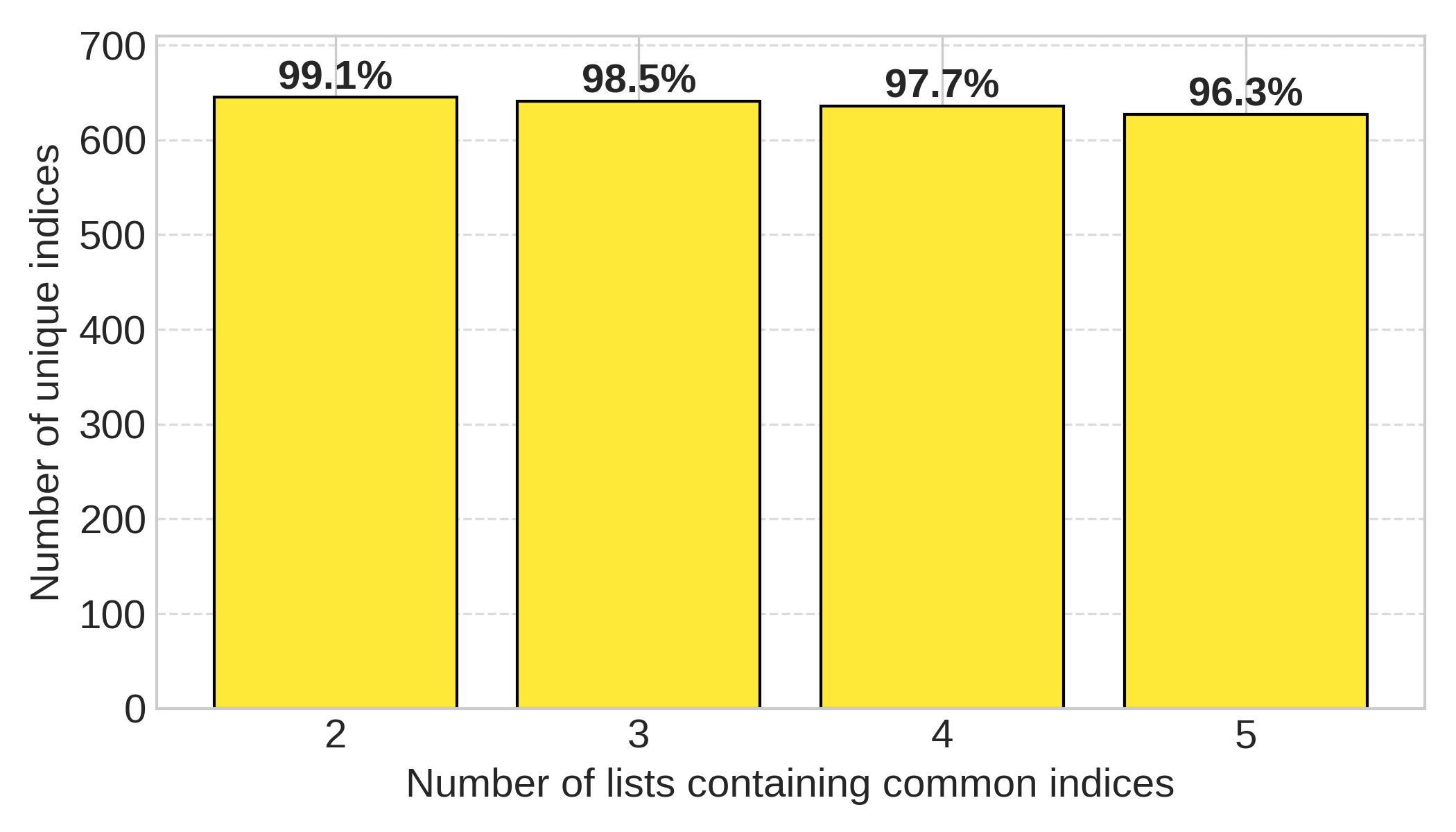}
        \captionsetup{skip=1pt}
        \label{fig:xcnn}
    \end{subfigure}
    \caption{Overlap of selected units across 5 random seeds on (Left) XSum dataset and (right) CNN/DailyMail dateset.}
    \label{fig:index_distributions}
\end{figure}

In Figure~\ref{fig:index_distributions}, we evaluate the stability and consistency of the selected units across five random seeds. As described in the experimental setup, for the CNN/DailyMail and XSum datasets, we randomly sampled 40 batches of examples to compute the task-relevant units. Thus, in this case, checking the consistency of unit selection remains important. As shown in the following figures, the 5 lists of indices obtained from the five random seeds exhibit a high overlap, with 91.5\% for XSum and 96.3\% for CNN/DailyMail, demonstrating that our unit selection strategy is both stable and consistent, even with a limited number of samples.

\subsubsection{Summarization on XSum and CNN/DailyMail: Compute Cost  } \label{compute cost}

We emphasize that Flex-KD computes gradients only with respect to the teacher’s final hidden layer, not through all layers. Consequently, the cost of importance estimation is independent of the teacher’s depth and remains lightweight even for large models. We emphasize also that we compute the gradient only one time before training the student. This one-time gradient computation requires a single backward pass over the dataset and takes only a few seconds on our tasks, as reported in Table 9 (on XSum and CNN datasets).
Nevertheless, in Table~\ref{tab:twenty}, we conduct an experiment to assess the compute overhead of Flex-KD on the summarization tasks, which involves selecting the task-relevant units for distillation. In this setting, the teacher model is BART-large with 406M parameters, and all experiments were run on a single A100 GPU. In our proposed method, units are selected only from the teacher’s last hidden layer. The results show that the additional gradient-based selection step introduces a negligible increase in runtime, around 26 seconds for XSum and 23 seconds for CNN/DailyMail, demonstrating that Flex-KD remains efficient in practice. 

During training, Flex-KD adds no substantial overhead: the feature-alignment loss operates solely on the selected subspace of the last layer, making its cost negligible relative to the forward/backward passes already required for logit KD. To address efficiency concerns directly, we additionally measured full training time (hours per epoch) for pure logit KD (KD) and for Projector, CKA-KD, and Flex-KD. As shown in Table~\ref{tab:fd_cost}, Flex-KD is the most efficient feature-distillation method, substantially faster than both Projector and CKA, because it avoids full-layer matching and performs alignment only at the final encoder/decoder layers. These results confirm that Flex-KD introduces minimal overhead while achieving significantly stronger performance.

\begin{table}[h!]
\centering
\caption{Compute overhead, results are averaged over 5 random seeds.}
\label{tab:twenty}
\begin{tabular}{lcc}
\toprule
& \textbf{XSum} & \textbf{CNN/DailyMail} \\
\midrule
time (seconds) & $25.98 \pm 1.27$ & $22.80 \pm 2.81$ \\
\bottomrule
\end{tabular}
\end{table}

\begin{table}[h!]
\centering
\caption{Computational cost (hours per epoch) of different feature distillation methods.
All experiments use a 6$\times$640 BART student and are run on a single A100 GPU.
Flex-KD applies feature alignment only on the last encoder and decoder layers (plus logit loss), while CKA and Projector perform full-layer alignment. On XSUM dataset.}
\label{tab:fd_cost}
{
\begin{tabular}{lcc}
\toprule
\textbf{Method} & \textbf{Hours / Epoch} \\
\midrule
KD \cite{hinton2015distilling} &  0.93\\
Projector \cite{jiao-etal-2020-tinybert} & 1.94 \\
CKA-KD \cite{dasguptaimproving} & 1.35 \\
\textbf{Flex-KD (ours)} & \textbf{0.93} \\
\bottomrule
\end{tabular}}
\end{table}

\section{Estimating functional contribution in practice} \label{aggregation}
To identify the teacher directions most relevant to downstream behavior, we compute functional contribution scores using gradients of the teacher’s output with respect to its final hidden representation. For an input sequence $x$ with $T$ valid tokens, let $H_T(x) \in \mathbb{R}^{T \times d_T}$ denote the teacher’s last-layer hidden states and $P_T(x,t)$ the corresponding token-level output distribution. We form a scalar functional
\[
\mathcal{L}(x) = \sum_{t=1}^{T} \sum_{v=1}^{V} \log P_T(x,t)_v,
\]
which aggregates log-probabilities across all tokens. Its gradient with respect to the hidden-state matrix yields
\[
\widetilde{G}(x)
    = \nabla_{H_T(x)} \mathcal{L}(x)
    \in \mathbb{R}^{T \times d_T},
\]
whose element-wise magnitude reflects local output sensitivity. Because functional relevance is defined across the entire sequence, we average these values over the token dimension to obtain a per-example importance vector
\[
g(x) = \frac{1}{T} \sum_{t=1}^{T} \big|\widetilde{G}(x)_{t,:}\big|
    \in \mathbb{R}^{d_T}.
\]
Aggregating over all training examples yields the dataset-level estimator
\[
\widehat{G}
    = \frac{1}{N} \sum_{j=1}^{N} g(x_j),
\]
which identifies the representation coordinates that consistently exhibit the largest functional contribution. The student-sized functional subspace is then defined by selecting the top-$d_S$ indices of $\widehat{G}$, which we use throughout all Flex-KD experiments.

Once $\widehat{G}$ is computed, we sort its coordinates in descending order and select the top-$d_S$ indices:
\[
E = \operatorname{TopK}(\widehat{G},\, d_S).
\]
These indices define a student-sized functional subspace consisting of the representation directions with the largest aggregate contribution to the teacher’s output. The corresponding restricted representation is
\[
h_T^{(E)}(x) \in \mathbb{R}^{d_S},
\]
obtained by retaining only the coordinates of $H_T(x)$ indexed by $E$.
\section{Activations Visualization}
\begin{figure}[h!]
\centering

% Row 1
\begin{subfigure}{0.37\linewidth}
    \includegraphics[width=\linewidth]{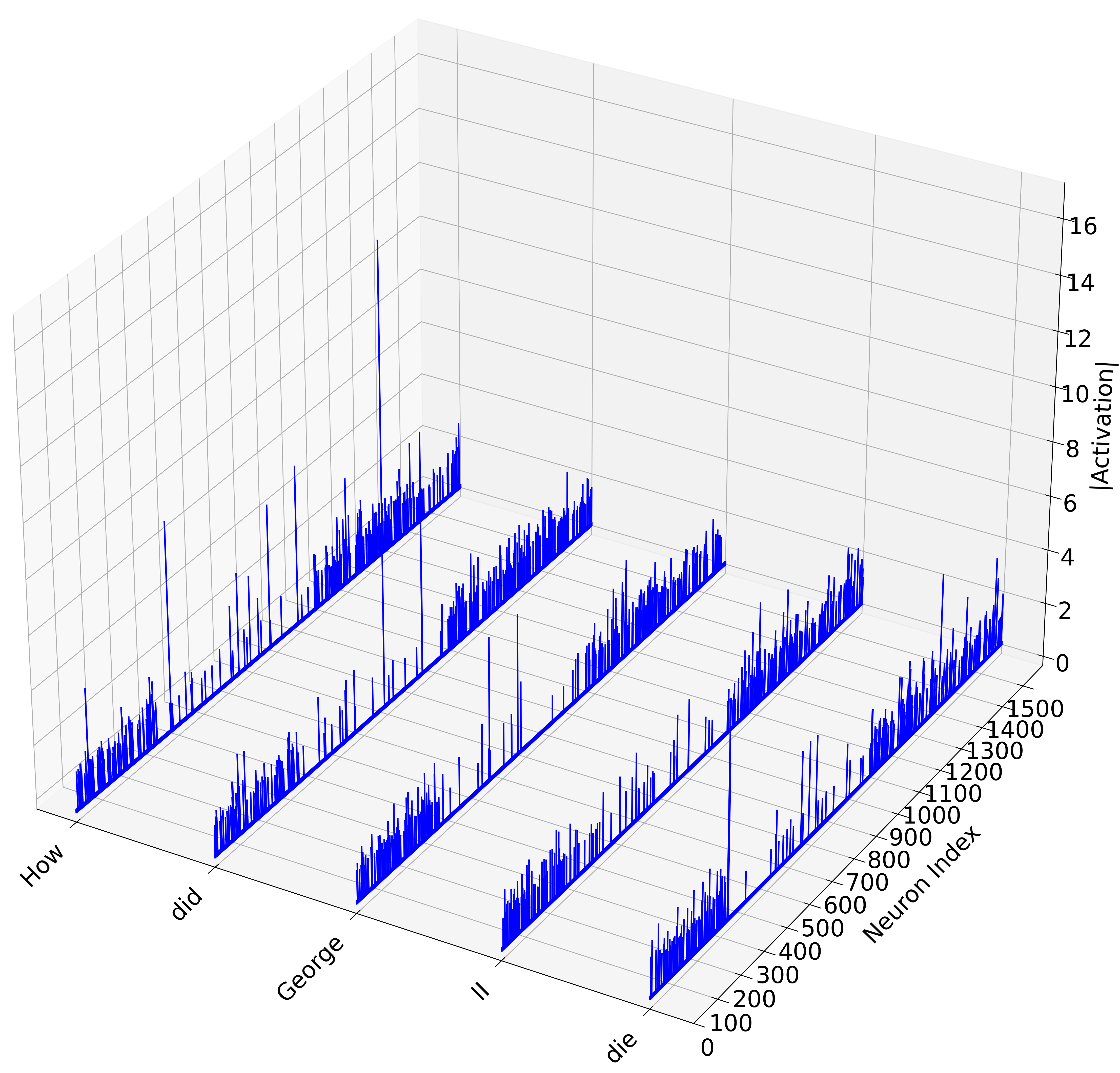}
    \caption{}
    \label{fig:george1}
\end{subfigure}
\hfill
\begin{subfigure}{0.37\linewidth}
    \includegraphics[width=\linewidth]{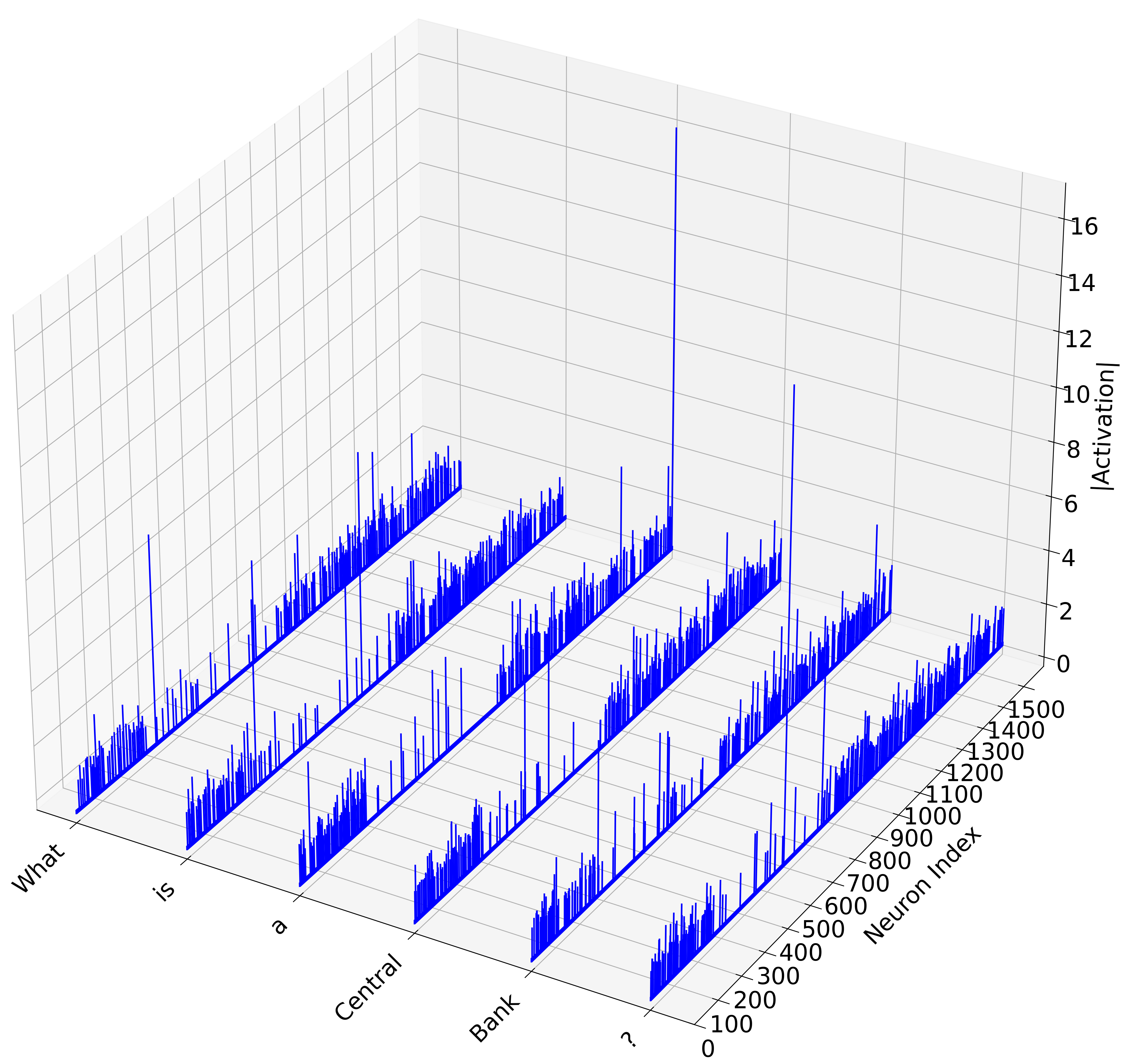}
    \caption{}
    \label{fig:bank1}
\end{subfigure}

\vskip\baselineskip

% Row 2
\begin{subfigure}{0.37\linewidth}
    \includegraphics[width=\linewidth]{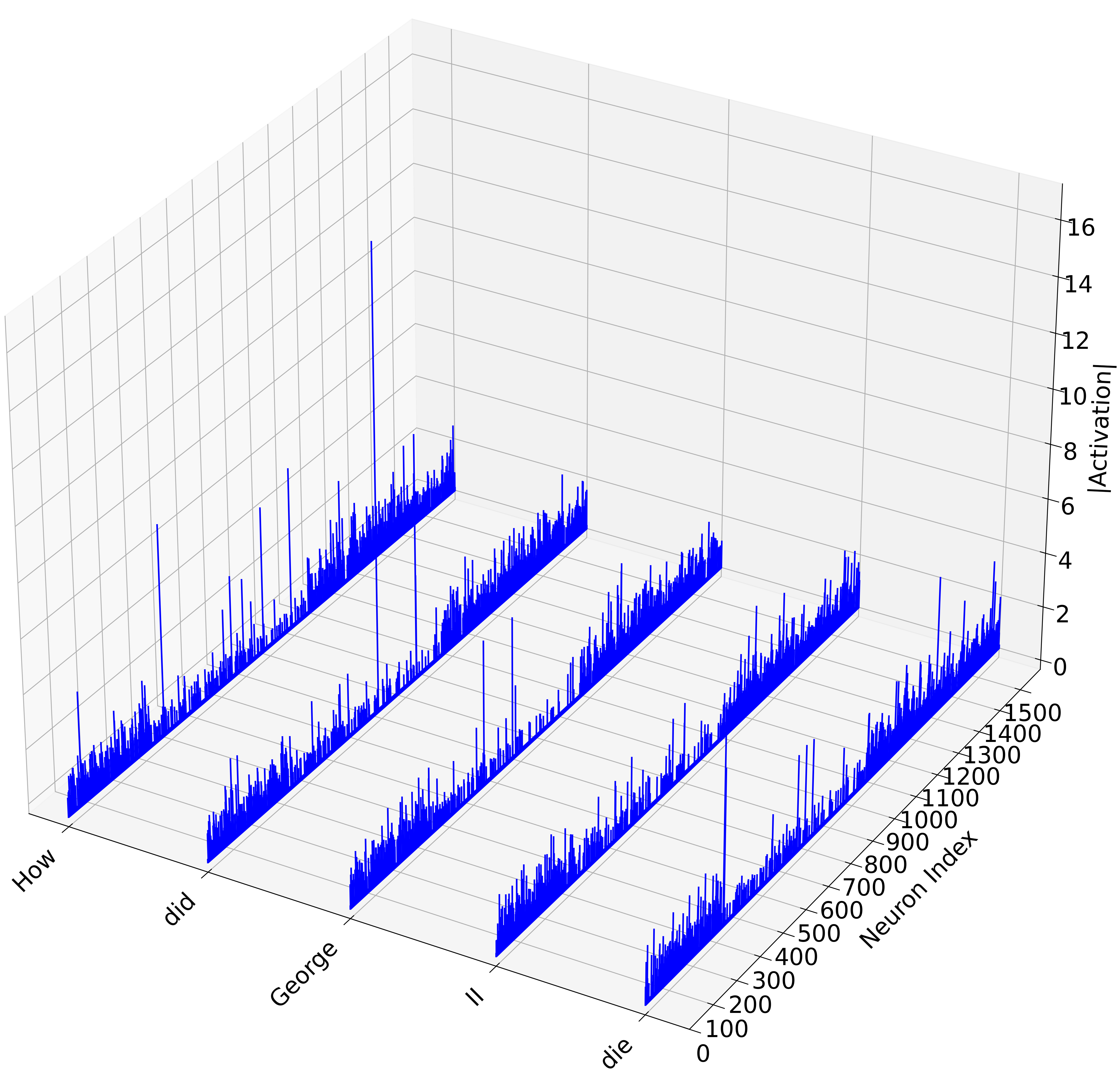}
    \caption{}
    \label{fig:george05}
\end{subfigure}
\hfill
\begin{subfigure}{0.37\linewidth}
    \includegraphics[width=\linewidth]{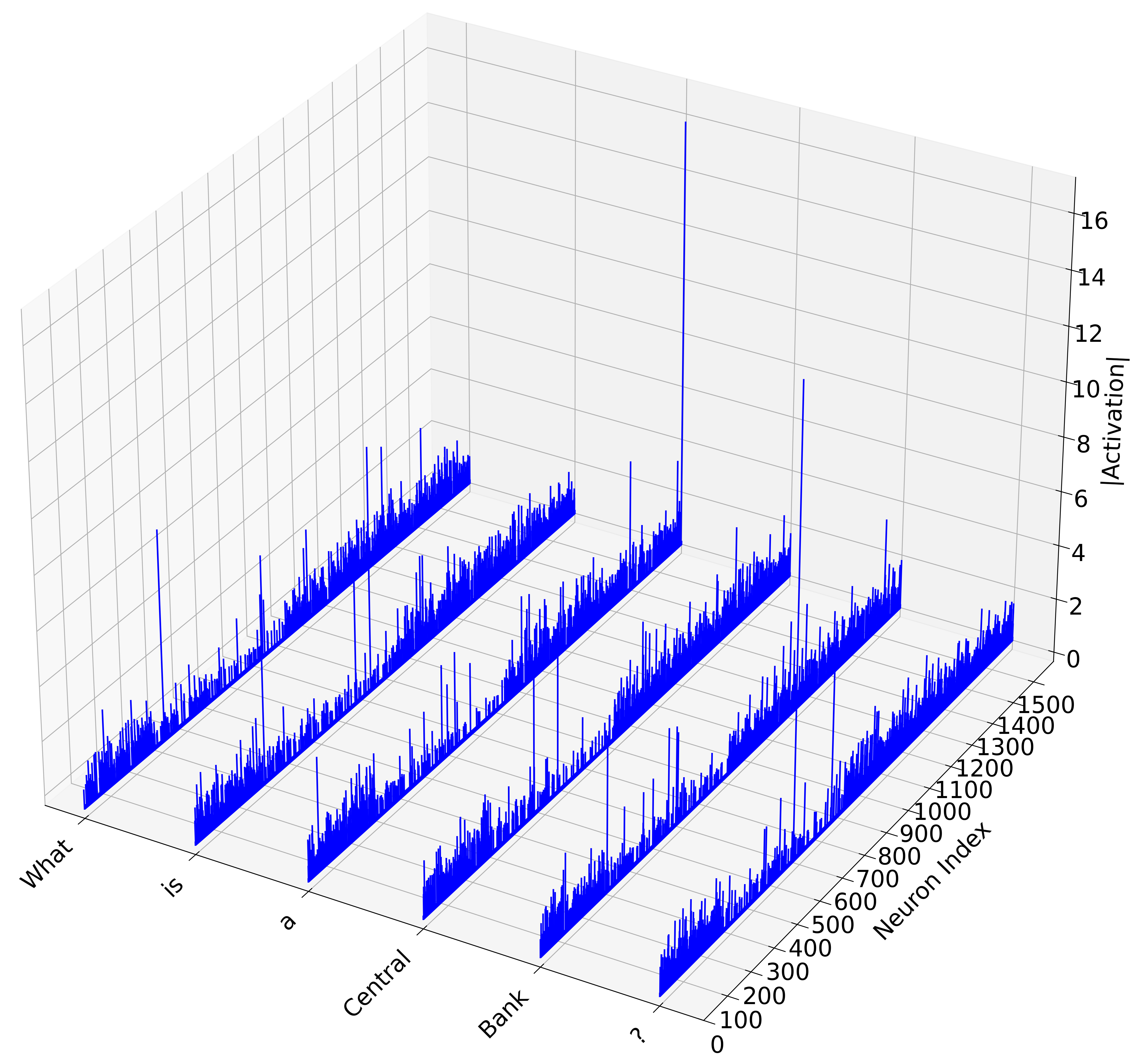}
    \caption{}
    \label{fig:bank05}
\end{subfigure}

\vskip\baselineskip

% Row 3 (single center-aligned image)
\begin{subfigure}{0.37\linewidth}
    \includegraphics[width=\linewidth]{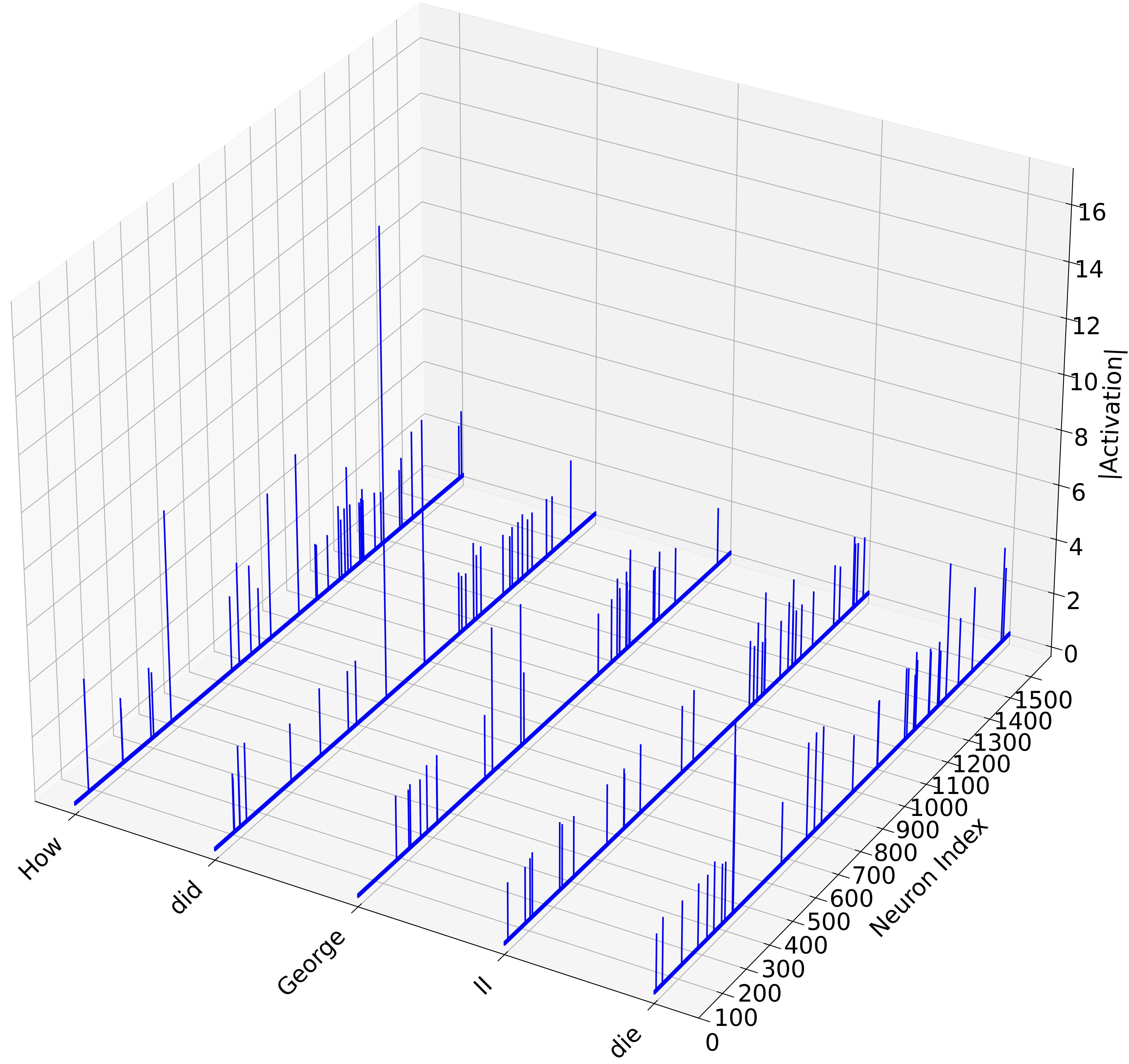}
    \caption{}
    \label{fig:george2}
\end{subfigure}
\hfill
\begin{subfigure}{0.37\linewidth}
    \includegraphics[width=\linewidth]{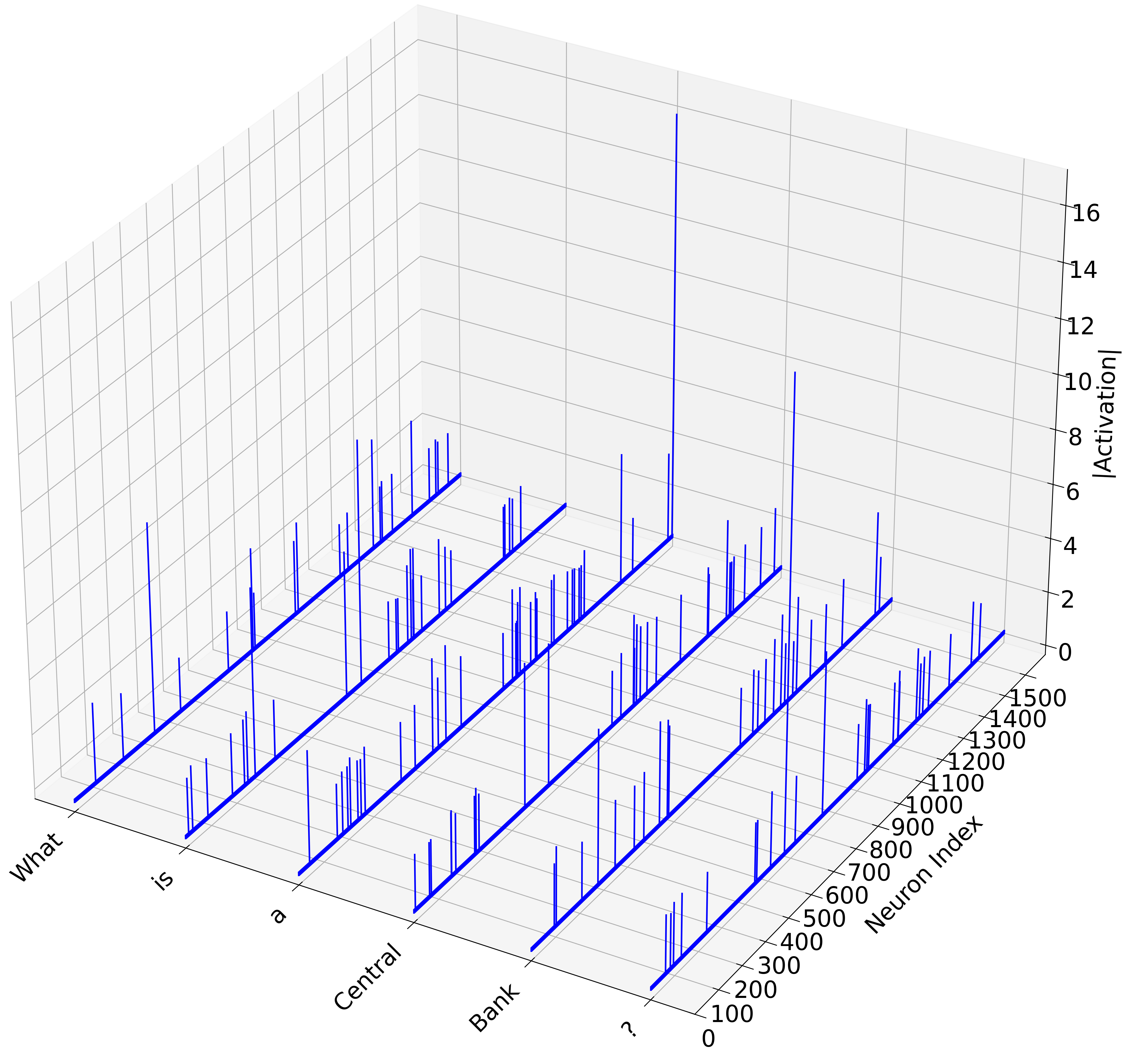}
    \caption{}
    \label{fig:bank2}
\end{subfigure}

\caption{Activation magnitudes (z-axis) after feeding training samples from the downstream task to a fine-tuned GPT-xlarge. x and y axes are sequence and feature dimensions, respectively: (a) and (b) We threshold values below
$1$ to zero. (c) and (d) We threshold values below
$0.5$ to zero. (e) and We (f) threshold values below
$2$ to zero.}
\label{fig:activation-all}
\end{figure}
In Figure~\ref{fig:activation-all}, we visualize the activations of the last hidden layer of a fine-tuned GPT-xlarge model on downstream examples. In~\ref{fig:george1} and~\ref{fig:bank1}, we threshold values $<1$ to $0$. In~\ref{fig:george05} and~\ref{fig:bank05}, we threshold values $<0.5$ to $0$. In~\ref{fig:george2} and~\ref{fig:bank2}, we threshold values $<2$ to $0$.  
As it can be seen, many units display near-zero or low-magnitude activations, suggesting limited contribution to the final output.

%%%%%%%%%%%%%%%%%%%%%%%%%%%%%%%%%%%%%%%%%%%%%%%%%%%%%%%%%%%%%%%%%%%%%%%%%%%%%%%
%%%%%%%%%%%%%%%%%%%%%%%%%%%%%%%%%%%%%%%%%%%%%%%%%%%%%%%%%%%%%%%%%%%%%%%%%%%%%%%

\end{document}